\documentclass[11pt, letterpaper, logo, onecolumn, copyright, numbering]{minimax}
\usepackage{etoolbox}

\usepackage[authoryear, sort&compress, round]{natbib}

\usepackage[inkscapeformat=png]{svg}

\usepackage[most]{tcolorbox}
\usepackage{academicons}

\usepackage{lipsum}
\usepackage{tabularx}
\usepackage{afterpage}
\usepackage{booktabs}
\usepackage{makecell}
\usepackage{multirow}
\usepackage{multicol} 
\usepackage{array}
\usepackage{float}
\usepackage{listings, listings-rust}
\usepackage{fontawesome5}
\usepackage{amssymb,graphicx}
\usepackage{subcaption}
\usepackage[dvipsnames]{xcolor}
\usepackage{hyperref}
\usepackage{cleveref}
\usepackage{longtable}
\usepackage{pdflscape}
\usepackage{adjustbox}
\usepackage{tikz}
\usetikzlibrary{calc,positioning,chains,shapes,arrows,fit,decorations.pathmorphing,patterns,fadings,shadows,patterns.meta}
\usepackage{wrapfig}
\usepackage{dialogue}
\usepackage{algorithm}
\usepackage{algorithmic}
\usepackage{colortbl}
\usepackage{mdframed}
\usepackage{fontawesome5}
\usepackage{colortbl}
\usepackage{amssymb}
\usepackage{textcomp}
\usepackage{marvosym}
\usepackage{listings} % 引入 code 包
\usepackage{xcolor}
\usepackage{CJKutf8}
\usepackage{tcolorbox} % Package for colored boxes
\usepackage[dvipsnames]{xcolor}
\usepackage{multicol}    % 引入 cabin 字体包
\usepackage{amsmath} 
\usepackage{cleveref}
\usepackage{hyperref}
\usepackage{graphicx}
\usepackage{multirow}

\theoremstyle{plain}

\theoremstyle{definition}

\theoremstyle{remark}

\usepackage{CJKutf8}

\usepackage{amsmath}
\usepackage[most]{tcolorbox}
\definecolor{darkgreen}{RGB}{0, 100, 0}
\usepackage{enumitem}
\usepackage[utf8]{inputenc} % allow utf-8 input
\usepackage[T1]{fontenc}    % use 8-bit T1 fonts
\usepackage{hyperref}       % hyperlinks
\usepackage{url}            % simple URL typesetting
\usepackage{booktabs}       % professional-quality tables
\usepackage{amsfonts}       % blackboard math symbols
\usepackage{nicefrac}       % compact symbols for 1/2, etc.
\usepackage{microtype}      % microtypography
\usepackage{xcolor}         % colors
\usepackage{colortbl}      % for \rowcolor
\usepackage[normalem]{ulem}
\usepackage{pifont}

\usepackage{fontawesome5}

\definecolor{medgray55}{gray}{0.55}
\definecolor{medgray}{gray}{0.7}
\definecolor{litegray}{gray}{0.9}
\definecolor{gblue}{RGB}{210, 227, 252}
\definecolor{gred}{RGB}{250, 210, 207}
\definecolor{gyellow}{RGB}{254, 239, 195}
\definecolor{ggreen}{RGB}{206, 234, 214}
\definecolor{gorange}{RGB}{254, 223, 200}

\definecolor{gblue9}{RGB}{23, 78, 166}
\definecolor{gred9}{RGB}{165, 14, 14}
\definecolor{gyellow9}{RGB}{227, 116, 0}
\definecolor{ggreen9}{RGB}{13, 101, 45}
\definecolor{gorange9}{RGB}{176, 96, 0}

\definecolor{myblue}{rgb}{0,0,1}
\definecolor{myred}{rgb}{1,0,0}
\definecolor{mylightgray}{gray}{0.95}

\definecolor{highlightblue}{HTML}{185ABC}

\makeatletter

\renewcommand\paragraph{\@startsection{paragraph}{4}{\z@}%
            {-2.5ex\@plus -1ex \@minus -.25ex}%
            {1.25ex \@plus .25ex}%
            {\itshape\normalsize\bfseries}}
\makeatother
\newcommand{\method}[1][WebExplorer]{\textsc{#1}}

\newcolumntype{L}[1]{>{\raggedright\let\newline\\\arraybackslash\hspace{0pt}}m{#1}}
\newcolumntype{C}[1]{>{\centering}m{#1}}

\newcolumntype{R}[1]{>{\raggedleft\let\newline\\\arraybackslash\hspace{0pt}}m{#1}}

\definecolor{ao}{rgb}{0.0, 0.0, 1.0}

\newcommand\vcent[1]{\vcenter{\hbox{#1}}}

\newcommand\loudspeaker[1][3]{\ensuremath{\vcent{\rule{.6ex}{.6ex}}\kern-.5ex%
  \vcent{\scalebox{.6}[1]{\rotatebox[origin=center]{90}{$\blacktriangle$}}}%
  \ifnum#1>0\relax\kern.05ex\vcent{\scalebox{.4}{\ttfamily)}}%
  \ifnum#1>1\relax\kern-.4ex\vcent{\scalebox{.56}{\ttfamily)}}%
  \ifnum#1>2\relax\kern-.55ex\vcent{\scalebox{.7}{\ttfamily)}}%
  \fi\fi\fi}%
}

\definecolor{green}{rgb}{0.9,0.9,0.9}

\crefname{figure}{Fig.}{Figs.}
\crefname{appendix}{Appx.}{Appx.}
\crefname{table}{Tab.}{Tables}
\Crefname{table}{Tab.}{Tables}
\crefname{section}{Sec.}{Sec.}
\Crefname{section}{Sec.}{Sec.}
\crefname{equation}{Eq.}{Eqs.}
\Crefname{equation}{Eq.}{Eqs.}
\crefformat{equation}{Eq.~#2#1#3}
\Crefformat{equation}{Eq.~#2#1#3}
\crefname{paragraph}{Sec.}{Secs.}

\definecolor{blue_light_1}{RGB}{221, 232, 255}
\definecolor{blue_1}{RGB}{50, 126, 230}

\usepackage{pifont}
\definecolor{uiLightBlue}{RGB}{187, 207, 234} % #BBCFEA

\newcommand{\circleS}{\raisebox{-0.1ex}{\textcolor{uiLightBlue!100}{\LARGE\ding{108}}}\kern-0.95em\raisebox{0.35ex}{\text{S}}}
\newcommand{\circleB}{\raisebox{-0.1ex}{\textcolor{orange!20}{\LARGE\ding{108}}}\kern-0.95em\raisebox{0.35ex}{\text{B}}}

\usepackage[svgnames]{xcolor}
\usepackage[most]{tcolorbox}
\usepackage{hyperref}
\definecolor{titleGreen}{RGB}{35,124,76}
\definecolor{titleText}{RGB}{255,255,255}
\definecolor{outerGreen}{RGB}{65,160,95} % 外框边框色
\definecolor{outerBg}{RGB}{245,255,245}  % 外框淡背景
\definecolor{stepOrange}{RGB}{242,146,23}
\definecolor{boxBg}{RGB}{255,249,235}
\definecolor{boxBorder}{RGB}{219,144,0}
\definecolor{thinkGray}{gray}{0.25}
\definecolor{thBg}{RGB}{248,248,255}   % GhostWhite
\definecolor{thFrame}{RGB}{102,102,170} % 紫蓝边
\definecolor{thTitle}{RGB}{70,70,140}

\definecolor{acBg}{RGB}{255,244,235}    % 浅橙底
\definecolor{acFrame}{RGB}{230,120,20}  % 橙边
\definecolor{acTitle}{RGB}{180,70,10}

\definecolor{obBg}{RGB}{240,255,255}    % 淡青底
\definecolor{obFrame}{RGB}{0,139,139}   % 深青边
\definecolor{obTitle}{RGB}{0,100,100}   % 深青标题

\definecolor{turnBlue}{RGB}{33,120,198}
\definecolor{turnBlueDark}{RGB}{22,86,150}
\definecolor{turnText}{RGB}{255,255,255}
\definecolor{turnOutline}{RGB}{60,120,160}

\tcbset{
  turnbase/.style={
    enhanced,
    left=8pt,right=8pt,top=5pt,bottom=5pt,
    fontupper=\bfseries\scshape,  % 粗体+小型大写
    halign=center,
  },
}

\newtcolorbox{turnbar}[1][]{%
  turnbase,
  colback=turnBlue,
  coltext=turnText,
  boxrule=0pt,
  arc=2pt,
  #1
}

\newtcolorbox{turnbargrad}[1][]{%
  turnbase,
  enhanced,
  interior style={
    left color=turnBlue,
    right color=turnBlueDark,
    middle color=turnBlue!85!turnBlueDark
  },
  coltext=turnText,
  boxrule=0pt,
  arc=2pt,
  #1
}

\newtcolorbox{turnbarline}[1][]{%
  turnbase,
  colback=white,
  coltext=turnBlueDark,
  colframe=turnOutline,
  boxrule=0.8pt,
  arc=2pt,
  #1
}

\tcbset{
mycallout/.style={
enhanced,
breakable,
rounded corners,
arc=8pt, outer arc=10pt,
boxrule=1.2pt,
colframe=black,
colback=blue!6!white,
left=10pt,right=10pt,top=8pt,bottom=12pt,
title={#1},
fonttitle=\bfseries\small\linespread{0.95}\selectfont,
coltitle=white,
colbacktitle=black,
attach boxed title to top center={yshift=-2pt},
boxed title style={
enhanced,
colframe=black,
colback=black,
boxrule=1.2pt,
arc=8pt,
left=10pt,right=10pt,
top=2pt,bottom=2pt
}
}
}

\newtcolorbox[use counter=mybox]{mycalloutbox}[1]{%
  enhanced,
  breakable,
  rounded corners,
  arc=8pt, outer arc=10pt,
  boxrule=1.2pt,
  colframe=black,
  colback=blue!6!white,
  left=10pt,right=10pt,top=8pt,bottom=12pt,
  title={#1},
  fonttitle=\bfseries\small\linespread{0.95}\selectfont,
  coltitle=white,
  colbacktitle=black,
  attach boxed title to top center={yshift=-2pt},
  boxed title style={
    enhanced,
    colframe=black,
    colback=black,
    boxrule=1.2pt,
    arc=8pt,
    left=10pt,right=10pt,
    top=2pt,bottom=2pt
  }
}
\tcbset{
  smallinner/.style={
    enhanced,
    arc=2pt,
    boxrule=0.7pt,
    left=7pt,right=7pt,top=7pt,bottom=7pt,
    before skip=6pt, after skip=6pt,
  },
  titlebar/.style={
    fonttitle=\bfseries,
    coltitle=white,
    attach boxed title to top left={yshift=-2pt, xshift=2pt},
    boxed title style={boxrule=0pt, colback=#1, arc=2pt, left=6pt,right=6pt,top=3pt,bottom=3pt},
  },
}
\newtcolorbox{outerframe}[1][]{
  colback=outerBg,
  colframe=outerGreen,
  boxrule=1pt,
  arc=4pt,
  left=10pt,right=10pt,top=10pt,bottom=10pt,
  enlarge left by=0mm,
  enlarge right by=0mm,
  enlarge top by=0mm,
  enlarge bottom by=0mm,
  #1
}

\newtcolorbox{topbanner}{
  colback=titleGreen,
  coltext=titleText,
  boxrule=0pt,
  left=10pt,right=10pt,top=7pt,bottom=7pt,
  fontupper=\Large\bfseries,
}

\newtcolorbox{stepbar}{
  colback=stepOrange,
  coltext=white,
  boxrule=0pt,
  left=6pt,right=6pt,top=4pt,bottom=4pt,
  fontupper=\bfseries,
  halign=center
}

\newtcolorbox{contentbox}[1][]{
  colback=boxBg,
  colframe=boxBorder,
  boxrule=0.8pt,
  left=10pt,right=10pt,top=10pt,bottom=10pt,
fontupper=\small,  % 添加这一行控制字体大小
  fonttitle=\small, 
  #1
}

\newtcolorbox{thoughtbox}[1][]{%
  smallinner,
  colback=thBg,
  colframe=thFrame,
  title=Thought,
  titlebar=thTitle,
  fontupper=\small,  % 添加这一行控制字体大小
  fonttitle=\small, 
  #1
}

\newtcolorbox{actionbox}[1][]{%
  smallinner,
  colback=acBg,
  colframe=acFrame,
  title=Action,
  titlebar=acTitle,
fontupper=\small,  % 添加这一行控制字体大小
  fonttitle=\small, 
  #1
}

\newtcolorbox{obbox}[1][]{%
  smallinner,
  colback=obBg,
  colframe=obFrame,
  title=Observation,
  titlebar=obTitle,
  fontupper=\small,  % 添加这一行控制字体大小
  fonttitle=\small, 
  #1
}

\makeatletter
\renewcommand\subparagraph{%
 \@startsection {subparagraph}{5}{\z@ }{3.25ex \@plus 1ex
 \@minus .2ex}{-1em}{\normalfont \normalsize \bfseries }}%
\makeatother

\let\cite\citep

\makeatletter % 允许修改内部命令
\renewcommand{\absfont}{\linespread{1.2}\fontsize{10}{12}\selectfont}
\makeatother  % 恢复 @ 的原始含义

\reportnumber{} % Leave blank if n/a

\renewcommand{\today}{}

\author{%
\centering
  \textbf{Junteng Liu}\textsuperscript{1*} \quad
  \textbf{Yunji Li}\textsuperscript{2*} \quad
  \textbf{Chi Zhang}\textsuperscript{2} \quad
  \textbf{Jingyang Li}\textsuperscript{2} \quad
  \textbf{Aili Chen}\textsuperscript{2} \quad
  \textbf{Ke Ji}\textsuperscript{2} \quad
  \textbf{Weiyu Cheng}\textsuperscript{2} \quad
  \textbf{Zijia Wu}\textsuperscript{2} \quad
  \textbf{Chengyu Du}\textsuperscript{2} \quad
  \textbf{Qidi Xu}\textsuperscript{2} \quad
  \textbf{Jiayuan Song}\textsuperscript{2} \quad
  \textbf{Zhengmao Zhu}\textsuperscript{2} \quad
  \textbf{Wenhu Chen}\textsuperscript{3} \quad
  \textbf{Pengyu Zhao}\textsuperscript{2} \quad
  \textbf{Junxian He}\textsuperscript{1} \\
  \textsuperscript{1}The Hong Kong University of Science and Technology \quad
  \textsuperscript{2}MiniMax \quad
  \textsuperscript{3}University of Waterloo \quad
  \texttt{jliugi@cse.ust.hk} \quad \texttt{junxianh@cse.ust.hk} \\
    \vspace{0.25em}
  \faGithub \quad  \texttt{\href{https://github.com/hkust-nlp/WebExplorer}{https://github.com/hkust-nlp/WebExplorer}} \\
}
\lstdefinestyle{noframe}{
  basicstyle=\footnotesize\ttfamily,
  breaklines=true,
  breakatwhitespace=true,
  columns=fullflexible,
  tabsize=2,
  showstringspaces=false,
  frame=none
}

\begin{abstract}

The paradigm of Large Language Models (LLMs) has increasingly shifted toward agentic applications, where web browsing capabilities are fundamental for retrieving information from diverse online sources. However, existing open-source web agents either demonstrate limited information-seeking abilities on complex tasks or lack transparent implementations. In this work, we identify that the key challenge lies in the scarcity of challenging data for information seeking. 
To address this limitation, we introduce \method{}: a systematic data generation approach using \emph{model-based exploration} and iterative, \emph{long-to-short query evolution}. This method creates challenging query-answer pairs that require multi-step reasoning and complex web navigation.
By leveraging our curated high-quality dataset, we successfully develop advanced web agent \method{}-8B through supervised fine-tuning followed by reinforcement learning. 
Our model supports 128K context length and up to 100 tool calling turns, enabling long-horizon problem solving. 
Across diverse information-seeking benchmarks, \method{}-8B achieves the state-of-the-art performance at its scale. Notably, as an 8B-sized model, \method{}-8B is able to effectively search over an average of 16 turns after RL training, achieving higher accuracy than WebSailor-72B on BrowseComp-en/zh and attaining the best performance among models up to 100B parameters on WebWalkerQA and FRAMES. Beyond these information-seeking tasks, our model also achieves strong generalization on the HLE benchmark even though it is only trained on knowledge-intensive QA data. These results highlight our approach as a practical path toward long-horizon web agents.
\end{abstract}

\begin{document}

\thispagestyle{empty}
% \vspace*{-2cm} % 向上调整位置
% \begin{center}
%   % \includegraphics[width=0.96\textwidth]{src/figures/title.pdf}
%   \includegraphics[width=1\textwidth]{src/figures/synLogic.jpg}
% \end{center}
% \vspace{-10pt}

% \title{\method: Implicit Graph Construction and Evolution for Training Long-Horizon Web Agents}
% \title{\method: Exploration-Driven Graph Construction and Query Evolution for Long-Horizon Web Agents}
% \title{\method: Model-Based Exploration and Evolution for Training Long-Horizon Web Agents}
\title{\method: Explore and Evolve for Training Long-Horizon Web Agents}
\maketitle
% \begin{center}
%   \includegraphics[width=\textwidth]{src/figures/title.pdf}
%   \vspace{-10pt}
% \end{center}
% \begin{figure}[t]
% \centering
%   \includegraphics[width=\textwidth]{src/figures/title.pdf}
%   \label{fig:banner}
%   \vspace{-10pt}
% \end{figure}
\begingroup
\renewcommand\thefootnote{\fnsymbol{footnote}}
% \footnotetext[1]{Equal contributions}
\endgroup

\begin{figure*} [h]
\begin{subfigure}[t]{0.325\textwidth} 
    \centering
    \includegraphics[width=\textwidth]{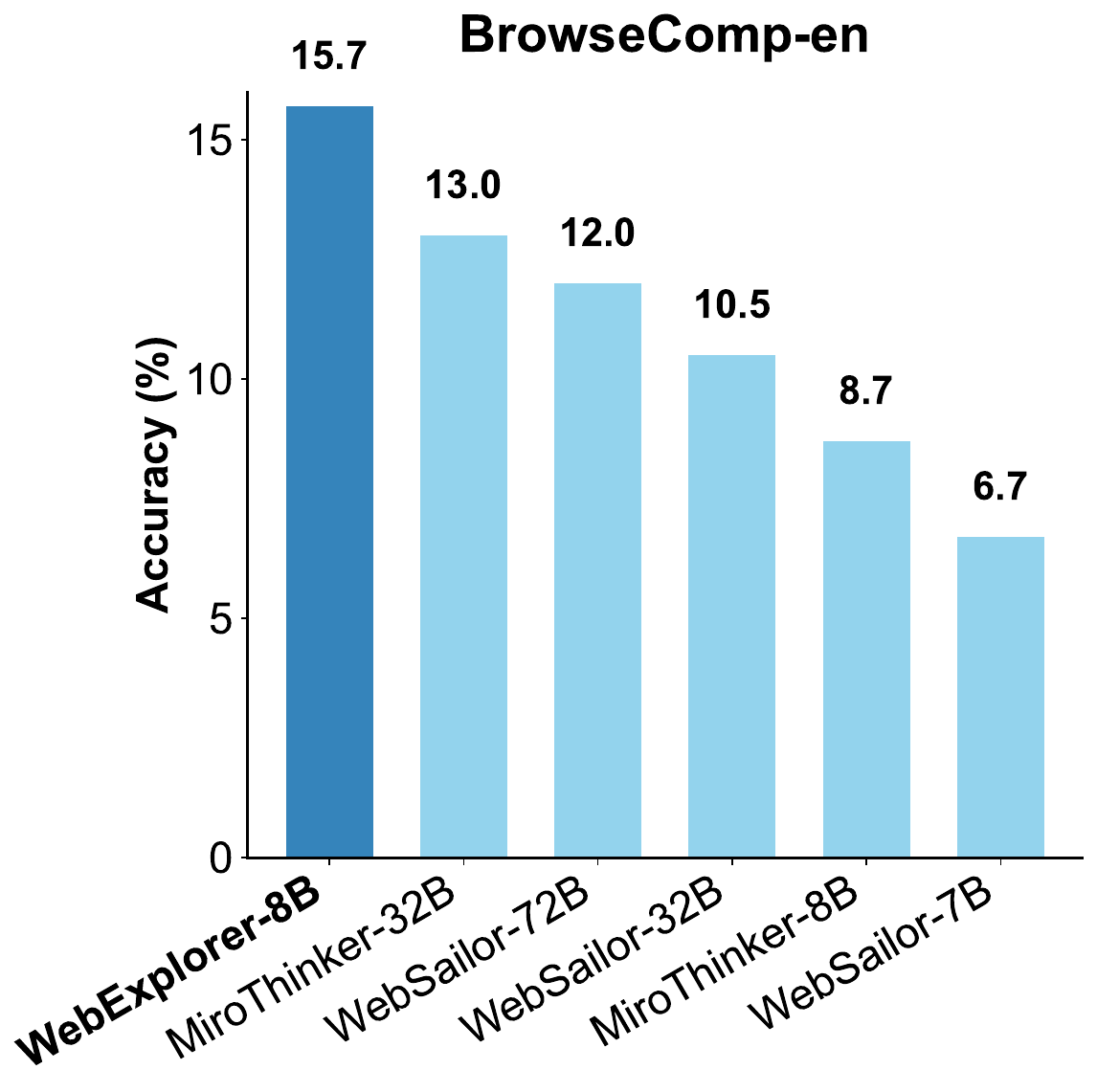}
\end{subfigure}
\hfill
\begin{subfigure}[t]{0.325\textwidth}
    \centering
    \includegraphics[width=\textwidth]{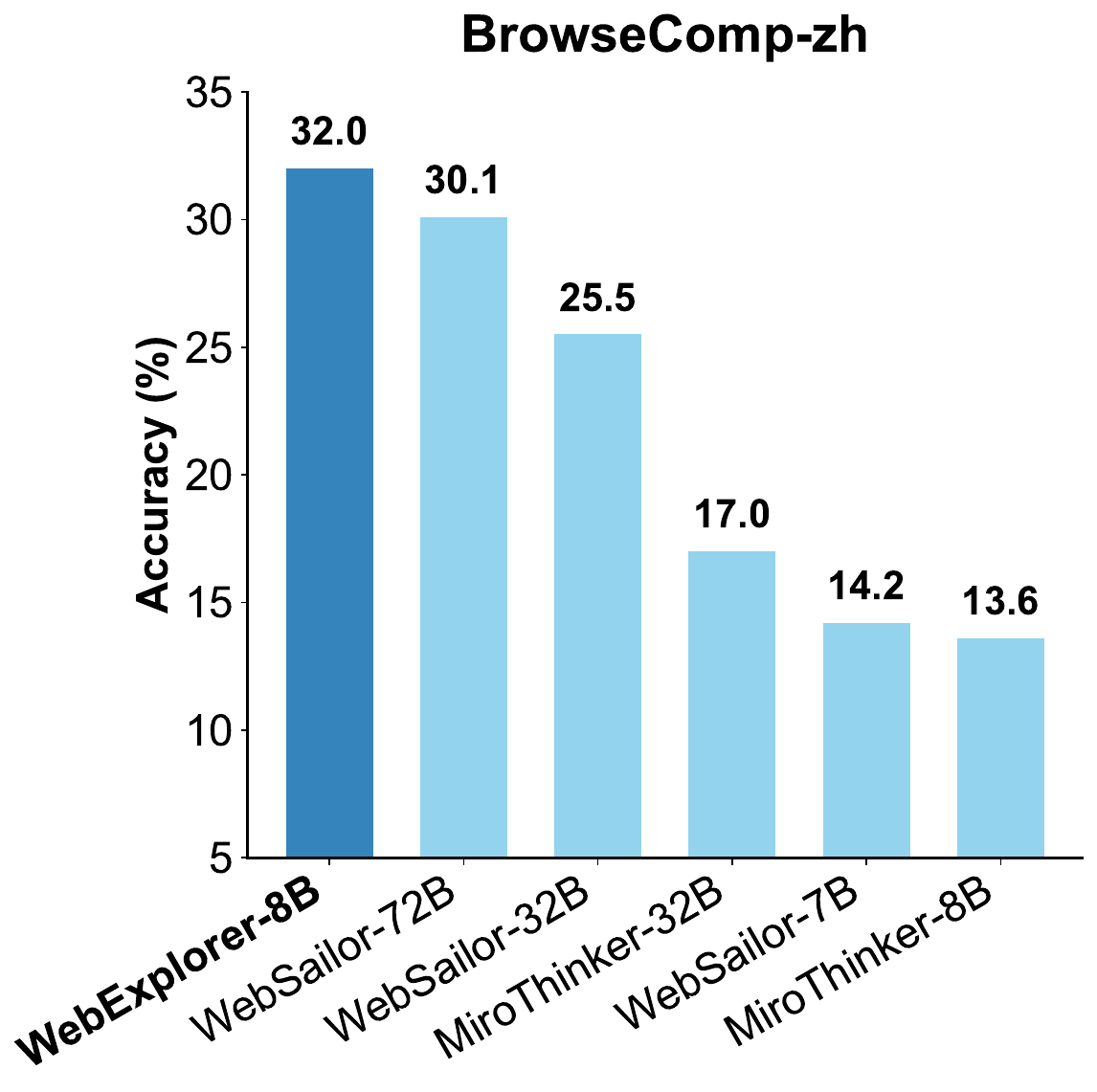}
\end{subfigure}
\hfill
\begin{subfigure}[t]{0.325\textwidth}
    \centering
    \includegraphics[width=\textwidth]{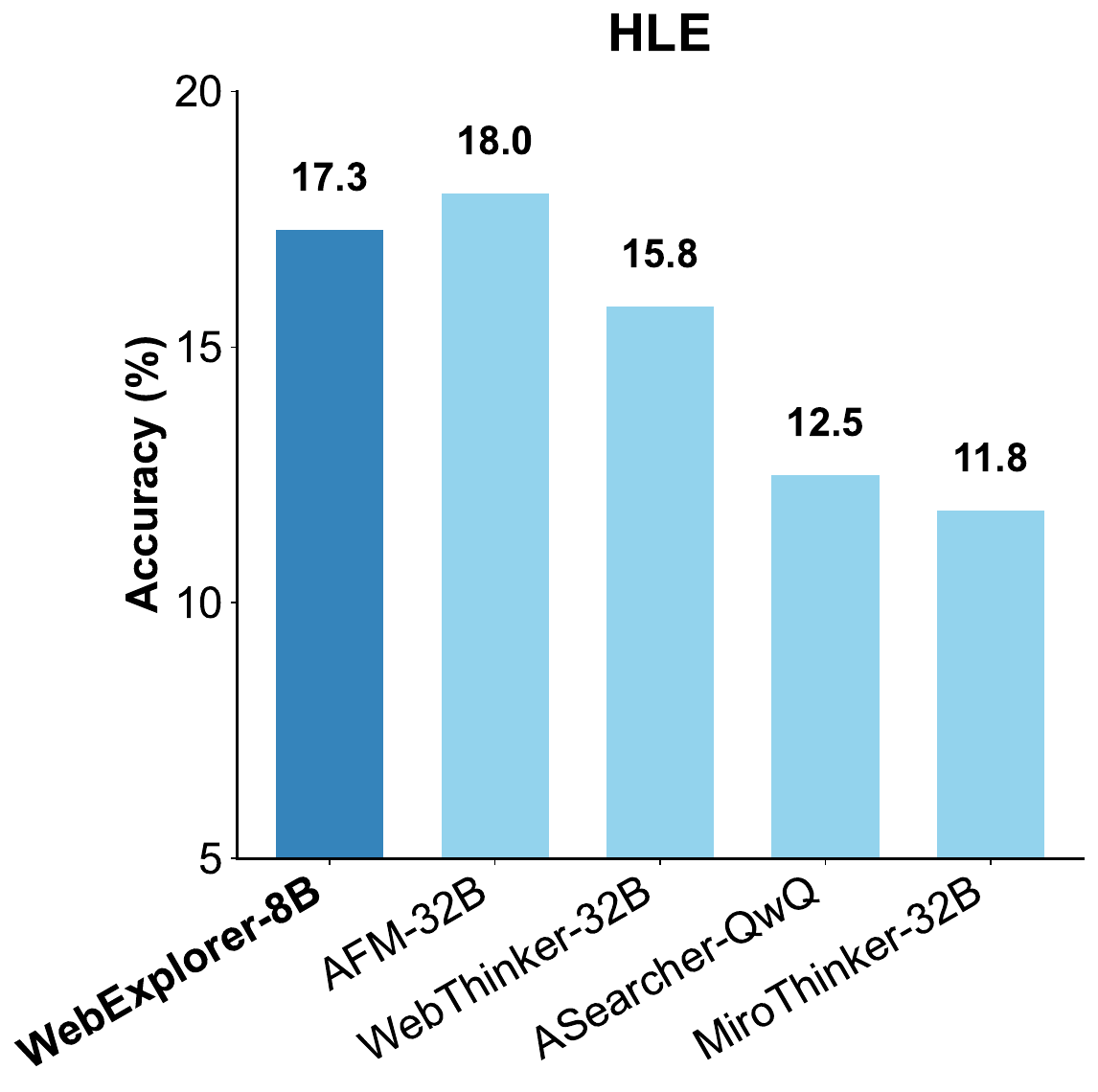}
\end{subfigure}
    \caption{Performance comparison on BrowseComp-en,  BrowseComp-zh and HLE benchmarks across different models.}

    \label{fig:perf-bars-flops}
\end{figure*}
% \begin{figure*}[h]
% \centering
% \includegraphics[width=0.325\textwidth]{src/figures/browsecomp_results.pdf}
% \hfill
% \includegraphics[width=0.325\textwidth]{src/figures/browsecomp_zh_results.pdf}
% \hfill
% \includegraphics[width=0.325\textwidth]{src/figures/hle_results.pdf}
% \caption{Performance comparison on BrowseComp-en, BrowseComp-zh and HLE benchmarks across different models.}
% \label{fig:perf-bars-flops}
% \end{figure*}
\section{Introduction}

The rapid advancement of Large Language Models (LLMs) has spurred a paradigm shift toward agentic intelligence, where models are equipped with diverse tools to tackle increasingly complex problems. Web browsing agents~\citep{openai_deep_research_2025,perplexity_deep_research_2025} represent a critical component of this shift, enabling autonomous information retrieval from the vast landscape of online sources. Despite significant progress, existing open-source models such as WebThinker~\citep{li2025webthinker} and WebSailor~\citep{li2025websailor} exhibit limited performance on challenging search-based benchmarks like BrowseComp-en/zh~\citep{wei2025browsecomp,zhou2025browsecomp}, BrowseComp-Plus~\citep{chen2025BrowseCompPlus} and  Xbench~\citep{xbench2025}, while stronger commercial models remain opaque in their training details~\citep{glm45_blog_2025,openai_deep_research_2025}.

We hypothesize that the fundamental challenge in developing capable web agents lies in the quality of training data. Current benchmarks have evolved to include queries of extraordinary difficulty -- For example, over half of BrowseComp-en queries fail to be solved by human annotators. While such challenging queries are rare in typical use cases, constructing high-quality, difficult query-answer pairs is essential for developing agents that can achieve super-human performance on information-seeking tasks.
Previous approaches to web navigation data construction have primarily employed two methodologies: Graph-based methods~\citep{li2025websailor, tao2025webshaper} construct explicit graphs through random walks where each node represents a website or entity and each edge represents a relationship, then utilize subgraphs to formulate QA pairs. Evolution-based approaches~\citep{wu2025webdancer,gao2025turnsunlockinglonghorizonagentic} iteratively modify existing queries to increase their complexity. 
However, these approaches face limitations. Previous graph construction introduces complexity in node expansion and selection heuristics. Meanwhile, evolution-based methods usually increase the complexity by injecting new information to replace original content at each evolution step, potentially resulting in unnatural query formulations and limited flexibility in synthesis. 
Our framework \method{} addresses these limitations through a novel approach that leverages model-based exploration to construct the information space and iterative long-to-short query evolution to reduce salient information. This enables more natural and flexible synthesis of challenging web navigation tasks.

Rather than constructing explicit graphs through rule-based methods, which is a complex process requiring decisions about expansion strategies and node selection, we adopt a simpler model-based approach to explore the information space. Starting from a seed entity, we leverage powerful LLMs to simulate the graph-building process internally through iterative search and browsing actions. This approach enables flexible, dynamic exploration of information spaces related to the seed entity without the overhead of explicit graph construction. The model then utilizes this explored information space to construct initial query-answer pairs.

% \jhc{Careful case studies revealed that this occurs because excessive information contained in queries provides clear search entry points that effectively bypass potential dead ends. In contrast, current challenging benchmarks like BrowseComp-en typically provide obfuscated information in queries and require several exploratory attempts to complete.}{}
% This observation suggests that the initial questions provide excessive information, making them easier to solve than desired.

In our preliminary experiments, however, we observed that these initially constructed QA pairs, while requiring multiple websites to solve, still remained relatively straightforward evidenced by the high success rates achieved by strong proprietary models. To address this limitation, we introduce a second-stage evolution process. 
% Given the excessive information provided in the initial QA pairs and inspired by the characteristics of current challenging benchmarks like BrowseComp-en, we employ an iterative evolution step. 
Specifically, unlike previous work where evolution involves injecting new information~\citep{wu2025webdancer,gao2025turnsunlockinglonghorizonagentic}, we prompt models to systematically increase query difficulty by \emph{removing} explicit clues and introducing strategic obfuscation. In contrast to the initial QA pairs with clear search entry points, the evolved QA pairs require longer solution processes with more exploratory search attempts. This systematic evolution generates challenging queries that lead to significantly lower success rates and require more reasoning steps from proprietary models. Through this process, we construct our \method{}-QA data.

Our training utilizes a typical approach combining supervised fine-tuning for cold-start initialization, followed by reinforcement learning using the GRPO algorithm~\citep{shao2024deepseekmath}. 
% \jhc{Leveraging high-quality trajectories from strong proprietary models for cold-start training, \method{}-8B (SFT) establishes strong search capabilities that already outperform the previous WebSailor-7B model}{}
% ~\citep{li2025websailor}.
Our RL training scales to 128K context length and 100 maximum number of tool calling turns, where we observe consistent increases in both the number of tool calls and benchmark performance. Based on Qwen3-8B~\citep{yang2025qwen3}, our \method{}-8B achieves state-of-the-art performance at its scale on multiple information-seeking benchmarks, including BrowseComp-en/zh~\citep{wei2025browsecomp,zhou2025browsecomp}, GAIA~\citep{mialon2024gaia}, WebWalkerQA~\citep{wu2025webwalker}, Frames~\citep{krishna2024frames}, and XBench-DeepSearch~\citep{xbench2025}. Notably, our model achieves 15.7\% on BrowseComp-en and 32.0\% on BrowseComp-zh, significantly outperforming the previous leading WebSailor 72B model despite a much smaller size. It also achieves 62.7\% on WebWalkerQA and 75.7\% on FRAMES, establishing the best performance among models up to 100B parameters. Beyond superior performance on information-seeking tasks, \method{}-8B demonstrates remarkable generalization to the academic benchmark HLE~\citep{phan2025humanity}, scoring 17.3\% and outperforming previous 32B models like WebThinker-32B~\citep{li2025webthinker}. This validates the strong generalization capability of our approach beyond pure information-seeking tasks.

\section{\method{}: Synthesizing Challenging QA Pairs}

\begin{figure*} [t]
    \centering
    \includegraphics[width=0.95\textwidth]
    % {images/dataset2.pdf}
    {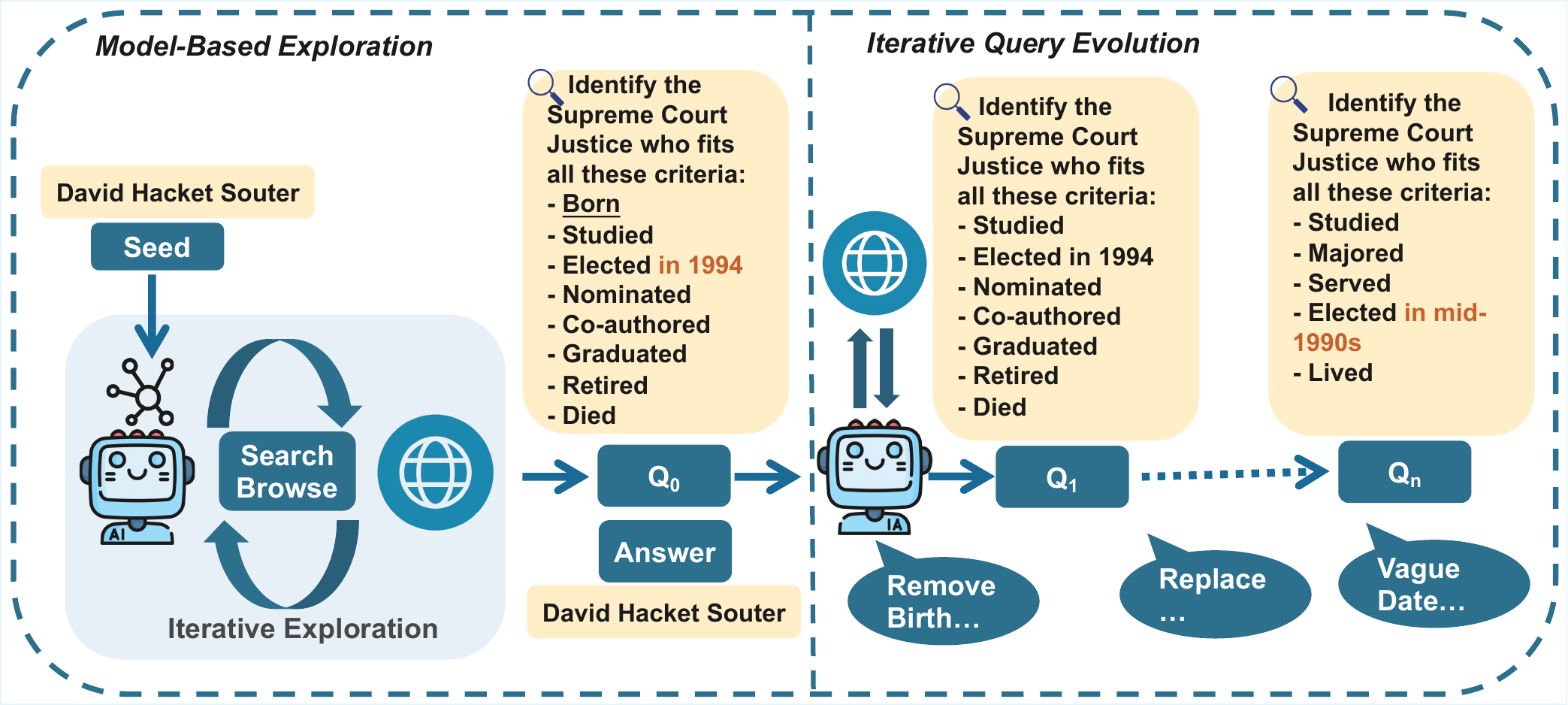}
    \caption{\textbf{Model-Based Exploration and Iterative Query Evolution Framework.} 
        Starting from a seed entity (e.g., David Hackett Souter), the framework employs iterative search and browsing actions to construct the information space related to the seed entity. Initial queries ($Q_0$) and Answers are generated based on this explored information space. Through iterative evolution, salient information is systematically obfuscated (e.g., ``Remove Birth...'', ``Replace ...'' or ``Vague Date...'') to produce more challenging queries ($Q_1$ to $Q_n$). This process ensures the resulting queries require longer reasoning steps and explorations.}
    \label{fig-pipeline}
\end{figure*}

Contemporary information-seeking benchmarks present great challenges, exemplified by BrowseComp-en~\citep{wei2025browsecomp} where more than half of the queries remain unsolvable by human annotators despite two hours of dedicated effort. These challenging benchmarks rely predominantly on manual curation~\citep{wei2025browsecomp,zhou2025browsecomp, mialon2024gaia}, resulting in prohibitively high annotation costs and severely limited dataset sizes that preclude their use in large-scale training. Concurrently, existing open-source training data exhibit some limitations: they either lack sufficient scale to enable large-scale model training or present overly simplistic queries that fail to match the difficulty of evaluation benchmarks~\citep{wu2025webdancer,li2025websailor,tao2025webshaper,pahuja-etal-2025-explorer}. Consequently, the autonomous synthesis of large-scale, challenging search-oriented query-answer pairs emerges as a pivotal research challenge in developing advanced web agents.

Several works have explored synthesizing challenging information-seeking QA pairs through two predominant methodological paradigms. Graph-based approaches explicitly construct knowledge graphs where each node represents a website or entity, then generate QA pairs by selecting subgraphs from the complete graph structure. Typically, larger subgraphs correspond to more challenging queries~\citep{li2025websailor,tao2025webshaper}. Evolution-based methods iteratively modify existing queries by injecting new information to systematically increase their complexity~\citep{wu2025webdancer,gao2025turnsunlockinglonghorizonagentic}, typically evolving queries to become progressively longer.
However, the previous rule-based graph construction introduces complexity of node expansion and selection heuristics. Instead, our method~\method{} utilizes existing LLMs to navigate the web and construct the information space autonomously. Additionally, traditional evolution approaches that inject new information at every step can become unnaturally complex. Rather than following short-to-long evolution, our iterative query evolution operates in a \emph{long-to-short} manner by strategically removing salient information, creating natural obfuscation.
As illustrated in Figure~\ref{fig-pipeline}, through these two stages, \method{} proposes a simple and flexible framework for challenging QA pair generation.
% As illustrated in Figure~\ref{fig-pipeline}, \jh{this sounds too trivial and superficial, like we just combine them together in an A+B manner. it is worth elaborating their limitations and motivate our approach better. For example, we can emphasize we utilize existing LLMs to navigate the web and construct the graph implicitly, rather than relying heuristic rules stuff (similar issues exist for the introduction, we don't want to introduce our approach as A+B)}\method{} integrates insights from both approaches and proposes a simple yet effective framework for challenging QA pair generation.

\subsection{Preliminaries}

Our agent framework builds upon the ReAct paradigm~\citep{yao2023react}, where agents execute iterative cycles of reasoning and acting. When presented with a query, the agent conducts multiple rounds of Thought-Action-Observation sequences. During each round, the language model formulates a Thought $\tau_t$ based on the current context, executes a structured Action $\alpha_t$, and receives corresponding environmental feedback as Observation $o_t$. The final round contains only Thought $\tau_t$ without further actions and observations.

Our action space encompasses two primary tools: $\alpha_t \in \{\texttt{search}, \texttt{browse}\}$. The \texttt{search}  action enables querying search engines to retrieve relevant information, while the \texttt{browse} action facilitates direct webpage access for content extraction. We formalize each action as $\alpha_t = (\alpha_t^m, \alpha_t^p)$, where $\alpha_t^m$ specifies the action type and $\alpha_t^p$ contains the necessary parameters. For \texttt{search} operations, $\alpha_t^p$ encompasses a list of query strings, whereas \texttt{browse} operations require target URLs and extraction objectives within $\alpha_t^p$. More details about these tools can be found in \textsection\ref{sec:exp-setup}.

The complete execution trajectory spanning $T$ iterations is represented as:
\begin{equation}
H_T = (\tau_0, \alpha_0, o_0, \tau_1, \alpha_1, o_1, \ldots, \tau_T)
\end{equation}

At each timestep $t$, the agent produces thought $\tau_t$ and selects action $\alpha_t$ according to policy $\pi(\tau_t, \alpha_t | H_{t-1})$, conditioning on the accumulated trajectory history $H_{t-1}$. We provide a detailed trajectory example in Appendix~\ref{app:traj-example}.

% \begin{figure*} [h]
%     % \vspace{-10pt}
%     \centering
%     \includegraphics[width=0.60\textwidth]{src/figures/synthesize_example_cropped.pdf}

% \caption{Illustration of implicit graph construction and initial Query-Answer pair synthesis.
% Starting from the seed ``Brazil National Team,'' the model iteratively explores using
% \circleS{} \enspace (\texttt{Search}) and
% \circleB{} \enspace (\texttt{Browse})
% actions to discover interconnected facts, then synthesizes a challenging query--answer pair that requires deep reasoning across multiple discovered connections.}
%     \label{fig-implicit}
% \end{figure*}

\begin{figure}[htbp]
\centering
\includegraphics[width=0.95\linewidth]{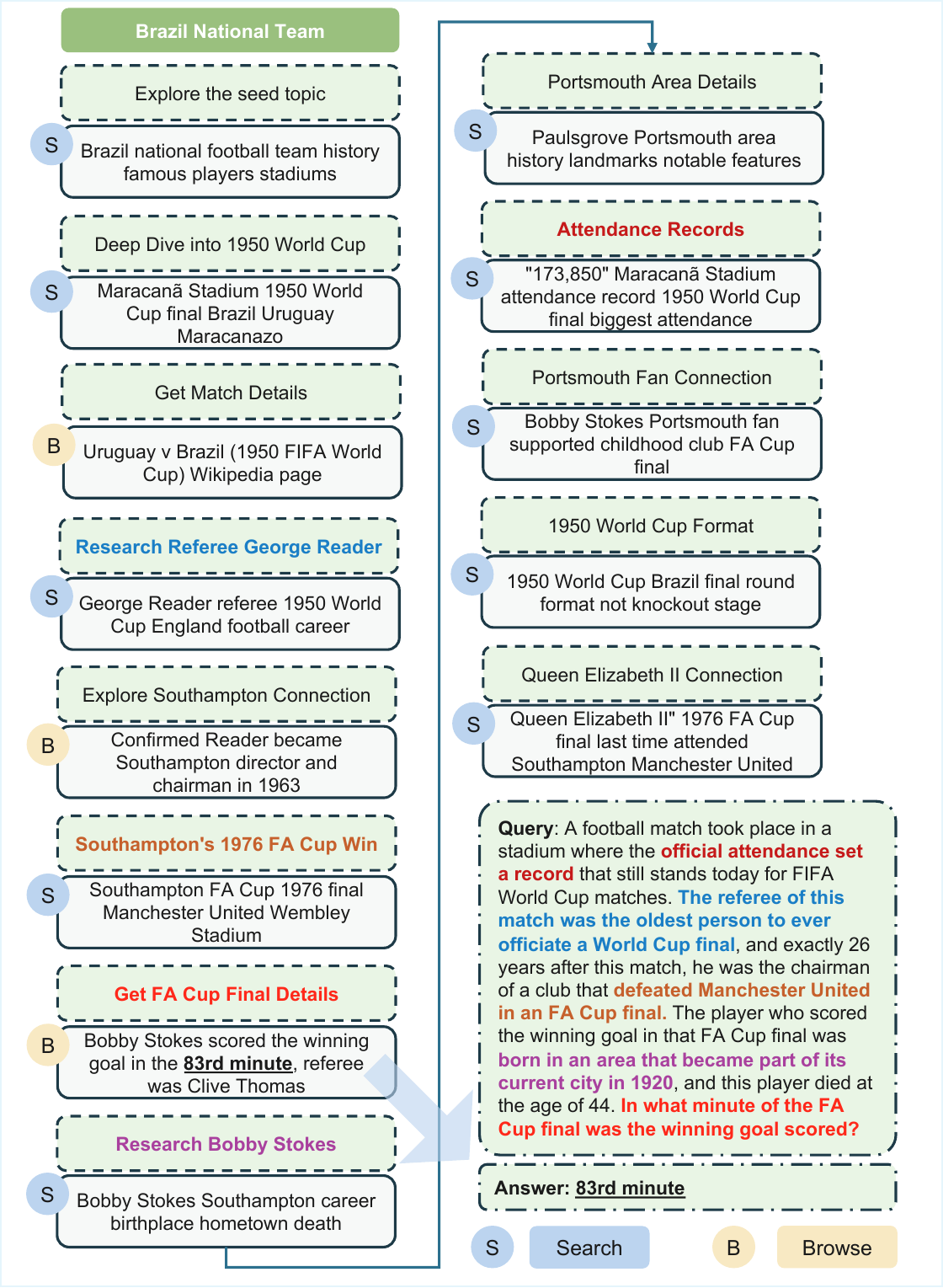}
\caption{Illustration of model-based exploration and initial Query-Answer pair synthesis.
Starting from the seed ``Brazil National Team'', the model iteratively explores using
\circleS{} \enspace (\texttt{Search}) and
\circleB{} \enspace (\texttt{Browse})
actions to discover interconnected facts, then synthesizes a challenging query--answer pair that requires deep reasoning across multiple discovered connections.}
\label{fig-implicit}
\end{figure}

% \subsection{Implicit Graph Construction via Model-Based Exploration}

\subsection{Model-Based Exploration}
Traditional graph-based methods ~\citep{li2025websailor,tao2025webshaper} require heuristic rules of explicit graph construction, typically initiated from a root URL or seed entity, followed by systematic expansion through predefined traversal strategies. 
The resulting graph contains structured information where usually nodes represent entities and edges form the relationships.
Such approaches involve iteratively identifying related entities, extracting their features, and expanding the graph until reaching predetermined size constraints. This process introduces some complexity, requiring careful, heuristic design of expansion strategies and node selection heuristics.

In contrast, we propose \method{}, which employs a different model-based exploration approach by leveraging powerful LLMs to construct the information space autonomously. 
Our method operates purely through prompting: we provide a seed entity as the initial search entry along with three example QA pairs, then instruct the model to conduct iterative search and browsing actions to explore the information space before synthesizing QA pairs.
Specifically, given an entity as a seed, the models conduct iterative search and browsing actions to construct a comprehensive information space encompassing entity-related content. This approach simulates the graph-building process internally, enabling flexible and dynamic exploration without the complexity of graph expansion strategy design and explicit graph maintenance. Subsequently, utilizing the explored information space, models generate QA pairs that necessitate reasoning across multiple websites to reach the correct solution.

As shown in Figure~\ref{fig-implicit}, starting from the seed ``Brazil National Team'', the model iteratively conducts \texttt{search} and \texttt{browse} actions to discover various interconnected facts. By focusing on one specific event—the ``1976 FA Cup''—the model explores diverse information including details about players and referees, ultimately constructing a Query-Answer pair that requires deep reasoning across multiple discovered connections.

Formally, starting from a seed entity $e_0$, the exploration process conducts multiple rounds of actions and observations:
\begin{equation}
H_T = (e_0,\tau_0, \alpha_0, o_0, \tau_1,  \alpha_1, o_1, \ldots, \tau_T)
\end{equation}

where $H$ represents the complete information space encompassing all explored content, and $\tau_T$ contains the synthesized QA pair. The model autonomously determines when to stop searching to formulate a challenging QA pair, eliminating the need for predefined stopping criteria and allowing the model to explore flexibly.
% \jh{when does the model decide to stop searching and synthesizing the query? is this a hyperparameter, or the model decides itself, plz explain}.

\subsection{Iterative Query Evolution}

\begin{mycalloutbox}{Example of BrowseComp-en}\label{box:browsecomp}
\textbf{Query:}
Please identify the fictional character who occasionally breaks the fourth wall with the audience, has a backstory involving help from selfless ascetics, is known for his humor, and had a TV show that aired between the 1960s and 1980s with fewer than 50 episodes.\quad\textit{\textbf{Answer:} Plastic Man}

\end{mycalloutbox}

% \usepackage{tcolorbox}
% While the initial QA pairs constructed from model-explored information spaces successfully incorporate content from multiple websites, we observe that proprietary models can still solve them with relatively high accuracy. This indicates that despite requiring multi-website reasoning, these initial QA pairs remain insufficiently challenging.

While the initial QA pairs constructed from model-explored information spaces successfully incorporate content from multiple websites, we observe that proprietary models can still solve them with relatively high accuracy. For instance, as shown in Table~\ref{tab:evolution_effect}, Claude-4-Sonnet achieves 86.6\% accuracy on the initial QA pairs compared to 12.2\% on BrowseComp-en and 68.3\% on GAIA. This indicates that despite requiring multi-website reasoning, these initial QA pairs remain insufficiently challenging. 
Through systematic case analysis, we identify that the initial queries contain excessive explicit information and salient clues that substantially reduce their difficulty. For example, the initial query-answer example shown below demonstrates several salient clues such as ``\textcolor{red}{\textit{the official attendance set a record}}'' and ``\textcolor{purple}{\textit{this player died at the age of 44}}'' that serve as clear and strong indicators. Such specific information—including dates, locations, and proper names—often provides direct entry points that enable straightforward solution trajectories without requiring exploratory detours or consideration of alternative reasoning paths.

In contrast, examination of challenging information-seeking benchmarks like BrowseComp-en reveals a critical distinguishing characteristic: these queries deliberately avoid providing clear, specific clues, instead employing vague descriptions. As the example above shows, the BrowseComp-en query ``\textit{TV show that aired between the 1960s and 1980s with fewer than 50 episodes}'' demonstrates this obfuscation. Therefore, these challenging queries usually do not contain clear search entry points, which necessitate extensive exploration and multiple reasoning attempts before reaching the correct answer.

Drawing inspiration from BrowseComp's design principles, we implement an iterative query evolution process to systematically increase QA pair difficulty. Unlike previous evolution methods that follow a short-to-long approach~\citep{wu2025webdancer,gao2025turnsunlockinglonghorizonagentic}, our evolution primarily reduces excessive information from initial QA pairs to increase difficulty. 
Given the complete initial QA construction information, we explicitly prompt models to refine queries through three strategic directions: (1) removing salient information, (2) introducing strategic obfuscation to specific details such as dates, locations, and proper names, and (3) identifying alternative descriptions to replace original explicit references. Through multiple iterative evolution cycles, we generate substantially more challenging QA pairs that require extensive exploration and demonstrate significantly reduced success rates.
As demonstrated in the example below, the evolved query eliminates some salient descriptions in \textcolor{purple}{purple} color like  \textcolor{purple}{\textit{this player died at the age of 44 }} and obscures concrete entities by replacing explicit references such as \textcolor{blue}{\textit{Manchester United}} with vague descriptors like \textcolor{blue}{\textit{First Division giant}}.

\begin{tcolorbox}[mycallout={Example of Evolved Query}]
\textbf{Initial Query:}
A football match took place in a stadium where \textcolor{red}{the official attendance set a record} that still stands today for FIFA World Cup matches. The referee of this match was \textcolor{purple}{the oldest person to ever officiate a World Cup final}, and \textcolor{purple}{exactly 26 years after this match}, he was the chairman of a club that defeated \textcolor{blue}{Manchester United} in an \textcolor{orange}{FA Cup final}. The player who scored the winning goal in that FA Cup final was born in an area that became part of its current city in \textcolor{gray}{1920}, \textcolor{purple}{and this player died at the age of 44}. In what minute of the FA Cup final was the winning goal scored? \quad\textit{\textbf{Answer:} 83rd minute}

\noindent\makebox[\linewidth]{\dotfill}

\textbf{Evolved Query:}
In the \textcolor{red}{unique} FIFA World Cup tournament format that concluded without a knockout final, a match official later guided a Second Division club to victory over a \textcolor{blue}{First Division giant} in the monarch's final attendance at such an \textcolor{orange}{occasion}. The match-winner had been rejected by the club he supported as a child, hailing from a district that joined a centuries-old Royal Naval stronghold \textcolor{gray}{two decades into the 20th century}. In which minute did this decisive strike occur?\quad\textit{\textbf{Answer:} 83rd minute}
\end{tcolorbox}

Formally, starting with the initial query $Q_0$ extracted from the previous information space $H^{(0)}$, we iteratively evolve queries through multiple rounds.  For $k = 0, \ldots, K-1$, the $(k{+}1)$-th information space $H^{(k+1)}$ is obtained by appending the $k$-th round’s augmentation to the entire history:

\begin{equation}
H^{(k+1)} = (H^k, \tau_1^{(k)}, \alpha_1^{(k)}, o_1^{(k)}, \ldots, \tau_{m_k}^{(k)}), \quad k = 0, \ldots, K-1
\end{equation}
% where each evolved query $Q_{k+1}$ is contained in $\tau_m^{(k)}$ and maintains the same answer $A$, and $(\alpha_1^{(k)}, o_1^{(k)}, \ldots, \alpha_{m-1}^{(k)}, o_{m-1}^{(k)})$ represents the multiple action-observation cycles conducted during the $k$-th evolution step to augment the information space.
where each evolved query $Q_{k+1}$ is contained in $\tau_{m_k}^{(k)}$ and maintains the same answer $A$. The sequence $(\alpha_1^{(k)}, o_1^{(k)}, \ldots, \alpha_{m_k-1}^{(k)}, o_{m_k-1}^{(k)})$ represents the multiple action-observation cycles conducted during the $k$-th evolution step to further augment the information space.

% \paragraph{Evolution as prefix extension.}
% Let $H^{(0)}$ be the initial trajectory. For $k = 0, \ldots, K-1$, the $(k{+}1)$-th information space is obtained by appending the $k$-th round’s augmentation to the entire history:
% \begin{equation}
% H^{(k+1)} \;=\; \bigl(H^{(k)}, \tau^{(k)}_1, \alpha^{(k)}_1, o^{(k)}_1, \ldots, \tau^{(k)}_{m_k}\bigr), \qquad k = 0, \ldots, K-1.
% \label{eq:evolve_prefix}
% \end{equation}
% Here, the evolved query $Q_{k+1}$ resides in $\tau^{(k)}_{m_k}$ and retains the same answer $A$.

\subsection{The \method{}-QA Dataset}
\label{sec:data}

% To ensure diversity in our synthesized QA pairs, we collect seed entities from Wikipedia and incorporate QA pairs from BrowseComp-en as exemplars in our synthesis prompt. After generating initial QA pairs using our model-based implicit graph construction, we apply our iterative query evolution methodology for 5 iterations to synthesize the \method{}-QA dataset. These evolved QA pairs, approximately 40K QA pair, are subsequently used for supervised fine-tuning and can directly facilitate reinforcement learning.
To ensure diversity in our synthesized QA pairs, we collect seed entities from Wikipedia and incorporate three QA pairs from BrowseComp-en as exemplars in our synthesis prompt. After generating initial QA pairs using our model-based exploration, we apply our iterative query evolution methodology for 5 iterations to synthesize the \method{}-QA dataset, with detailed prompts description for the synthesis shown in Appendix~\ref{app-prompt-details}. These evolved final QA pairs, approximately 40K in total, are subsequently used for supervised fine-tuning and can be used directly for reinforcement learning.

To validate the quality of \method{}-QA and demonstrate the effectiveness of our iterative evolution process, we conduct a comprehensive comparative analysis across multiple web navigation datasets. Our evaluation encompasses our initial QA pairs, the final evolved QA pairs, and established datasets from prior work, including WebDancer~\citep{wu2025webdancer}, SailorFog~\citep{li2025websailor}, WebShaper~\citep{tao2025webshaper}, and ASearcher~\citep{gao2025turnsunlockinglonghorizonagentic}. Using Claude-4-Sonnet as our evaluation model, we assess both accuracy and the average number of tool calling turns required to solve each QA pair. Additionally, we employ a more powerful model, OpenAI-o3, to analyze the tool calling turns distribution between our Initial QA and Evolved QA, as well as between Evolved QA and the challenging benchmark BrowseComp-en.

\begin{table}[t]
\centering
\setlength{\tabcolsep}{4pt}
\begin{tabular}{l|cc|cccc}
\toprule
\textbf{Dataset} & \textbf{Initial QA} & \textbf{Evolved QA} & \textbf{WebDancer} & \textbf{SailorFog} & \textbf{WebShaper} & \textbf{ASearcher} \\
\midrule
Average Turns & 7.9 & \textbf{9.9} & 5.4 & 8.2 & 8.4 & 6.5 \\
Accuracy (\%) & 86.6 & 67.1 & 62.0 & 35.0 & 67.4 & 62.0 \\
\bottomrule
\end{tabular}
\caption{Comparison of average tool calling turns and accuracy (\%) of Claude-4-Sonnet across web navigation datasets, reporting our Initial QA, Evolved QA, and other QA datasets  (WebDancer, SailorFog, WebShaper, ASearcher). }
\label{tab:evolution_effect}
\end{table}

\begin{figure*}[!t]
\begin{subfigure}[t]{0.49\textwidth}
    \centering
    \includegraphics[width=\textwidth]{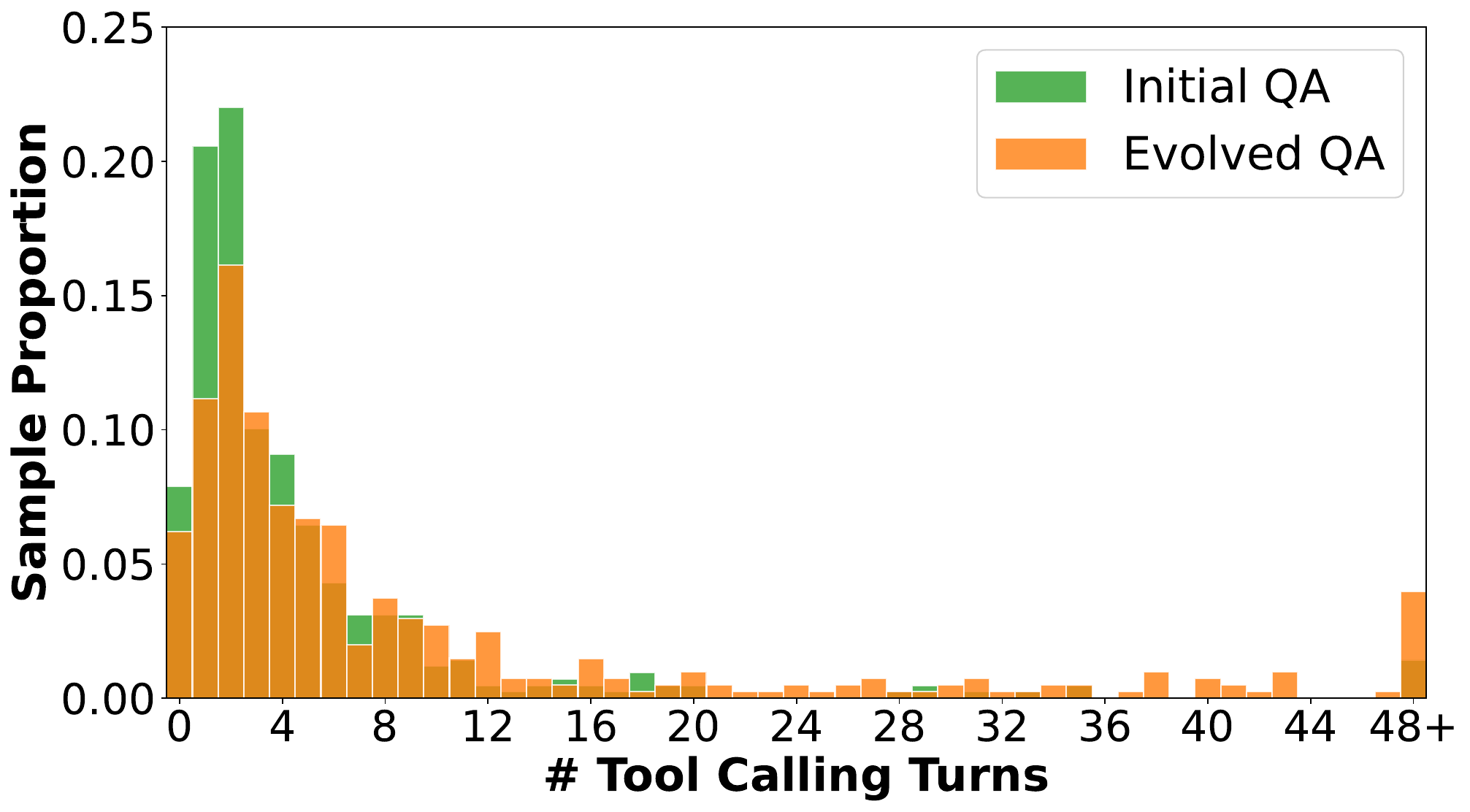}
    \caption{Initial QA vs Evolved QA}
    \label{fig:initial-evolved-comparison}
\end{subfigure}
\hfill
\begin{subfigure}[t]{0.49\textwidth}
    \centering
    \includegraphics[width=\textwidth]{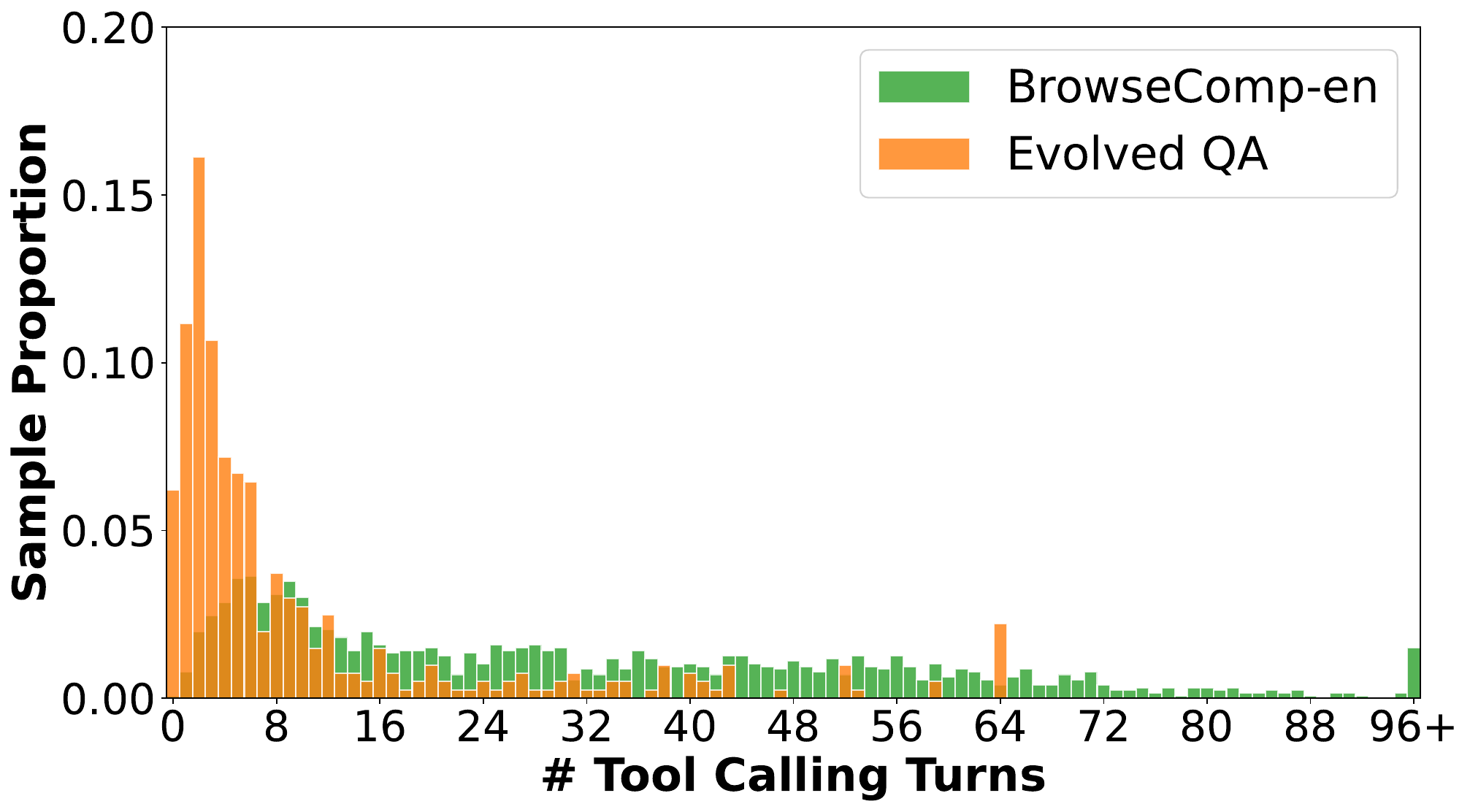}
    \caption{Evolved QA vs BrowseComp-en}
    \label{fig:evolved-browsecomp-comparison}
\end{subfigure}
\caption{Tool calling turns distribution comparisons using OpenAI-o3: Initial QA vs Evolved QA (left) and Evolved QA vs BrowseComp-en (right).}
\label{fig:tool-distribution}
\end{figure*}

The results presented in Table~\ref{tab:evolution_effect} demonstrate the critical importance of our iterative evolution process for generating challenging QA pairs. The evolution methodology significantly transforms the initial queries, with accuracy dropping substantially from 86.6\% to 67.1\%, while the average number of solution turns increases notably from 7.9 to 9.9. This substantial shift indicates that our evolution process successfully creates more complex queries requiring extensive multi-step reasoning and exploration.
Furthermore, when compared to existing datasets from other works, our evolved \method{}-QA achieves the highest average turn count, demonstrating the superior complexity and quality of our synthesized data. 
The detailed tool calling distribution shown in Figure~\ref{fig:tool-distribution} (left) validates the effectiveness of our evolution process, where easy QA pairs solvable within 4 turns decrease significantly. While Figure~\ref{fig:tool-distribution} (right) shows that a gap remains between our Evolved QA and BrowseComp-en in terms of tool calling turns, BrowseComp-en presents excessive difficulty with accuracy below 20\% on most open-source models. Therefore, completely mirroring BrowseComp-en's difficulty level is unnecessary for training current open-source models.

Overall, \method{}-QA, this increased complexity, while maintaining reasonable solvability, establishes our dataset as high-quality training data that effectively advances web navigation agent capabilities.

\section{Cold Start and Reinforcement learning}
% Our training approach follows the standard post-training two-phase paradigm: supervised fine-tuning for cold start initialization, followed by reinforcement learning for advanced capability development~\citep{guo2025deepseek,glm45_blog_2025}. The initial supervised fine-tuning phase enables models to learn proper invocation of search and browse functions while developing foundational long-horizon search capabilities. Subsequently, reinforcement learning further scales reasoning abilities, extending to longer contexts and increased maximum turn limits.

Our training methodology adopts the established post-training two-phase paradigm: supervised fine-tuning for cold start initialization, followed by reinforcement learning for advanced capability development~\citep{guo2025deepseek,glm45_blog_2025}. The initial supervised fine-tuning phase enables models to acquire proper invocation of search and browse functions while developing foundational long-horizon search capabilities. Subsequently, reinforcement learning further enhances reasoning abilities, extending model performance to longer contexts and increased maximum turn limits, ultimately achieving more advanced long-horizon problem-solving behaviors.

\subsection{Supervised Fine-tuning for Cold Start}

After synthesizing the challenging QA pairs as described in ~\textsection\ref{sec:data}, we leverage the commercial model to collect high-quality trajectories for supervised fine-tuning. We employ rejection sampling during data collection, ensuring that our fine-tuning process exclusively utilizes correct trajectories.
We adopt the ReAct framework~\citep{yao2023react} as our foundational format, incorporating search and browse as the two primary tools for actions $\alpha$, enclosed by \texttt{<tool\_call>} and \texttt{</tool\_call>} tags. The framework includes explicit reasoning thoughts $\tau$ denoted by \texttt{<think>} tags, enabling transparent cognitive processes, and tool responses as observations $o$ marked by \texttt{<tool\_response>} tags. Our collected trajectories encompass multiple rounds of action $\alpha$, thought $\tau$, and observation $o$ sequences, with an example demonstrated in Appendix~\ref{app:traj-example}.

During this phase, the model learns to decompose complex queries into manageable sub-problems, execute appropriate tool calls, interpret returned information, and synthesize findings into coherent responses. This cold start initialization is crucial for establishing the fundamental capabilities required for subsequent reinforcement learning optimization.

\subsection{Reinforcement Learning}
Following the cold-start phase, which endows the model with fundamental search and browse action capabilities along with long-horizon reasoning ability, we conduct further reinforcement learning training to enhance reasoning performance and optimize decision-making strategies using GRPO algorithm~\citep{shao2024deepseekmath}.  Notably, in the RL phase, we can directly utilize the synthesized QA pairs without requiring trajectory demonstrations for the questions, allowing the model to explore diverse reasoning paths autonomously.
For reward design, we implement a composite reward function that balances structural correctness with answer accuracy:
\begin{equation}
R = 0.2 \cdot R_{\text{format}} + R_{\text{correct}}
\end{equation}

Format rewards $R_{\text{format}}$ evaluate the correctness of response formatting, primarily assessing whether tool calls and thought structures adhere to the specified format requirements. This component ensures that the model maintains proper interaction protocols with external tools and preserves the interpretability of its reasoning process. For accuracy rewards $R_{\text{correct}}$, we leverage the DeepSeek-V3 model~\citep{liu2024deepseek} as an automated judge to evaluate whether the final responses are correct given the ground truth answers. This automated evaluation approach enables scalable assessment while maintaining high reliability.

To accommodate the model's expanding reasoning capabilities during training, we implement a progressive context length expansion strategy. We begin with a maximum length of 64K tokens and a tool calling turn limit of 50. As the model generates increasingly complex trajectories, we gradually increase the maximum length to 96K tokens with a 75-turn limit, and finally to 128K tokens with a 100-turn limit. This progressive expansion allows the model to develop more sophisticated long-horizon reasoning patterns throughout the training process.

\section{Experiments}

\begin{table}[t]
\centering
\small
\setlength{\tabcolsep}{3.5pt}
\begin{tabular}{lccccccccc}
\toprule
\textbf{Model}  & \textbf{BC-en} & \textbf{BC-zh} & \textbf{GAIA} & \textbf{WebWalkerQA} & \textbf{FRAMES} & \textbf{Xbench-DS}  & \textbf{HLE} \\
\midrule
OpenAI-o3$^{\dagger}$ & 50.9 & 58.1 & 70.5$^{\dagger}$ & 71.7 & 84.0 & 66.7 & 20.2 \\
Claude-4-Sonnet$^{\dagger}$  & 12.2 & 29.1 & 68.3$^{\dagger}$ & 61.7 & 80.7 & 64.6  & 20.3 \\
\midrule
GLM-4.5 &  26.4 & 37.5 & 66.0$^{\dagger}$ & 65.6$^{\dagger}$ & 78.9$^{\dagger}$ & 70.0$^{\dagger}$ & 21.2$^{\dagger}$ \\
DeepSeek-V3.1 & 30.0 & 49.2 & 63.1$^{\dagger}$ & 61.2$^{\dagger}$ & 83.7&  71.2 & 29.8   \\
Kimi-K2$^{\dagger}$ & 14.1 &28.8 & 57.7 & 63.0 & 72.0& 50.0 & 18.1 \\
\midrule\midrule
WebShaper-72B  & - & - & \textbf{60.0} & 52.2 & - & - & - \\
WebShaper-32B (QwQ)  & - & - & 53.3 & 49.7 & - & - & - \\
WebShaper-32B  & - & - & 52.4 & 51.4 & - & - & - \\
\midrule
WebSailor-72B  & 12.0 & 30.1 & 55.4 & - & - & \textbf{55.0} & - \\
WebSailor-32B  & 10.5 & 25.5 & 53.2 & - & - & 53.3 & - \\
WebSailor-7B  & 6.7 & 14.2 & 33.0 & - & - & 34.3 & - \\
\midrule
ASearcher-Web-QwQ  & 5.2 & 15.6 & 52.8 & 34.3 & 70.9 & 42.1 & 12.5 \\
\midrule
AFM-32B &  11.1 & - & 55.3 & \textbf{63.0} & - & - & \textbf{18.0}\\
AFM-7B  & 5.8 & -  & 40.8 & 55.6 & - & - & 15.6  \\
\midrule
WebThinker-32B & 2.8 & - & 48.5 & 46.5 & - & - & 15.8 \\
\midrule
MiroThinker-32B-DPO-v0.1  & 13.0 & 17.0 & 57.3 & 49.3 & 71.7 & - & 11.8 \\
MiroThinker-8B-DPO-v0.1  & 8.7 & 13.6 & 46.6 & 45.7 & 64.4 & - & - \\
\midrule
\rowcolor{blue!10} \method{}-8B (SFT) & 7.9 & 21.3 & 43.7 & 59.8 & 72.6 & 47.5 & 16.0 \\
\rowcolor{blue!10} \method{}-8B (RL) & \underline{\textbf{15.7}} & \underline{\textbf{32.0}} & \underline{50.0} & {\underline{62.7}} & \underline{\textbf{75.7}} & \underline{53.7} & {\underline{17.3}} \\
\bottomrule
\end{tabular}
\caption{Accuracy (\%) of web agents on information-seeking benchmarks. BC-en and BC-zh denote BrowseComp-en and BrowseComp-zh respectively. XBench-DS refers to XBench-DeepSearch. \textbf{Bold} indicates the best performance among open-source models $<$ 100B, while \underline{underlined} values represent the best performance among models $<$ 10B parameters. All scores of \method{}-8B are computed as Avg@4 using LLM-as-Judge. Entries marked with a dagger ($\dagger$) were reproduced by us under our scaffold: on model name = entire row; on a number = that entry only.}
\label{tab:model_comparison}
\end{table}

% \begin{figure*} [t]
%     \centering
%     \includegraphics[width=0.5\textwidth]
%     % {images/dataset2.pdf}
%     {src/figures/hle_analysis.pdf}

%     \caption{}

%     % \vspace{-10pt}
%     \label{fig-hle}
% \end{figure*}

\subsection{Experimental Setups}
\label{sec:exp-setup}

\noindent \textbf{Benchmarks} 
We employ several information-seeking benchmarks to comprehensively evaluate different web agent abilities. These include BrowseComp-en~\citep{wei2025browsecomp}, one of the most challenging information-seeking benchmarks introduced by OpenAI, and its Chinese counterpart BrowseComp-zh~\citep{zhou2025browsecomp}. We also evaluate on GAIA~\citep{mialon2024gaia}, a widely-adopted benchmark for General AI Assistants, where we use the 103-sample text-only validation subset following WebShaper. Additional benchmarks include WebWalkerQA~\citep{wu2025webwalker} for systematic data extraction across website subpages, FRAMES~\citep{krishna2024frames} for testing factual responses and reasoning capabilities, and XBench-DeepSearch~\citep{xbench2025} focusing on deep information retrieval tasks.
Beyond these information-seeking benchmarks, we also evaluate on the frontier academic benchmark HLE~\citep{phan2025humanity} to verify our model's generalization capabilities -- which contains many extremely challenging STEM questions, while our training data is all about Wikipedia-alike knowledge. We report all scores on these benchmarks using Avg@4 and employ LLM-as-Judge evaluation~\citep{liu2023calibrating} with DeepSeek-V3~\citep{liu2024deepseek} following previous work~\citep{li2025websailor,tao2025webshaper}.

% , and ScholarSearch~\citep{zhou2025scholars} for academic information retrieval.

\noindent \textbf{Models} We compare our approach against both proprietary and open-source agents. For proprietary models, we primarily benchmark against OpenAI-o3 and  Claude-4-Sonnet~\citep{claude4_blog_2025}. 
Among open-source agents, we compare against advanced models including, GLM-4.5~\citep{glm45_blog_2025}, DeepSeek-V3.1~\citep{liu2024deepseek}, Kimi-K2~\citep{team2025kimi}, WebShaper~\citep{tao2025webshaper}, WebSailor~\citep{li2025websailor}, ASearcher~\citep{gao2025turnsunlockinglonghorizonagentic}, AFM~\citep{li2025chainofagentsendtoendagentfoundation}, WebThinker~\citep{li2025webthinker} and MiroThinker~\citep{miromind2025odr}.

% \noindent \textbf{Scaffold Details} Following previous work~\citep{li2025websailor,tao2025webshaper}, our agent scaffold contains two tools: \texttt{search} and \texttt{browse}. The \texttt{search} tool takes one or multiple queries and returns the top-10 Google search results with title, URL, and snippet for each query. The \texttt{browse} tool takes a URL and a browse query, then uses Jina~\citep{jina2025} to retrieve the full content and employs Gemini 2.5 Flash~\citep{comanici2025gemini} to answer the browse query based on the URL content. Using this scaffold, we evaluate OpenAI o3, Claude-4-Sonnet, and Kimi-K2 models, ensuring fair comparison across all models. However, there is an exception, we have to note that MiroThinker models utilize more tools than search and browse.

\noindent \textbf{Scaffold Details} Following previous work~\citep{li2025websailor,tao2025webshaper}, our agent scaffold contains two tools: \texttt{search} and \texttt{browse}. The \texttt{search} tool takes one or multiple queries and returns the top-10 Google search results with title, URL, and snippet for each query. The \texttt{browse} tool takes a URL and a browse query, then uses Jina~\citep{jina2025} to retrieve the full content and employs Gemini 2.5 Flash~\citep{comanici2025gemini} to answer the browse query based on the URL content. Further details about these two tools can be found in Appendix~\ref{sec:tool-schemas}. 
Using this unified scaffold, we evaluate OpenAI-o3, Claude-4-Sonnet, and Kimi-K2 on all benchmarks, and evaluate GLM-4.5 and DeepSeek-V3.1 on a subset (marked with $\dagger$ where the original reports lack results), ensuring a fair cross-model comparison. Scaffolds of most other baselines generally rely on search and browse tools, making comparisons fair; the exception is MiroThinker, which employs additional tools beyond \texttt{search} and \texttt{browse}.
% Note that scores for MiroThinker models are taken from their reported results, where they utilize additional tools beyond search and browse, which may provide them with certain advantages in the evaluation.

\noindent \textbf{Training Details} 
We use the Qwen3-8B model~\citep{yang2025qwen3} to conduct SFT and RL training, resulting in the \method{}-8B model. For supervised fine-tuning, we utilize approximately 13K training samples with a training batch size of 32 and a learning rate of 1e-5 for 4 epochs. For reinforcement learning, we train on approximately 12K samples using the GRPO algorithm, where each group consists of 8 rollouts with a batch size of 64 and a learning rate of 1e-6. During the RL process, we gradually increase the maximum response length to 128K and the maximum turns limit to 100.

\begin{figure*} [t]
\begin{subfigure}[t]{0.32\textwidth}  % 注意这里用 [t]
    \centering
    \includegraphics[width=\textwidth]{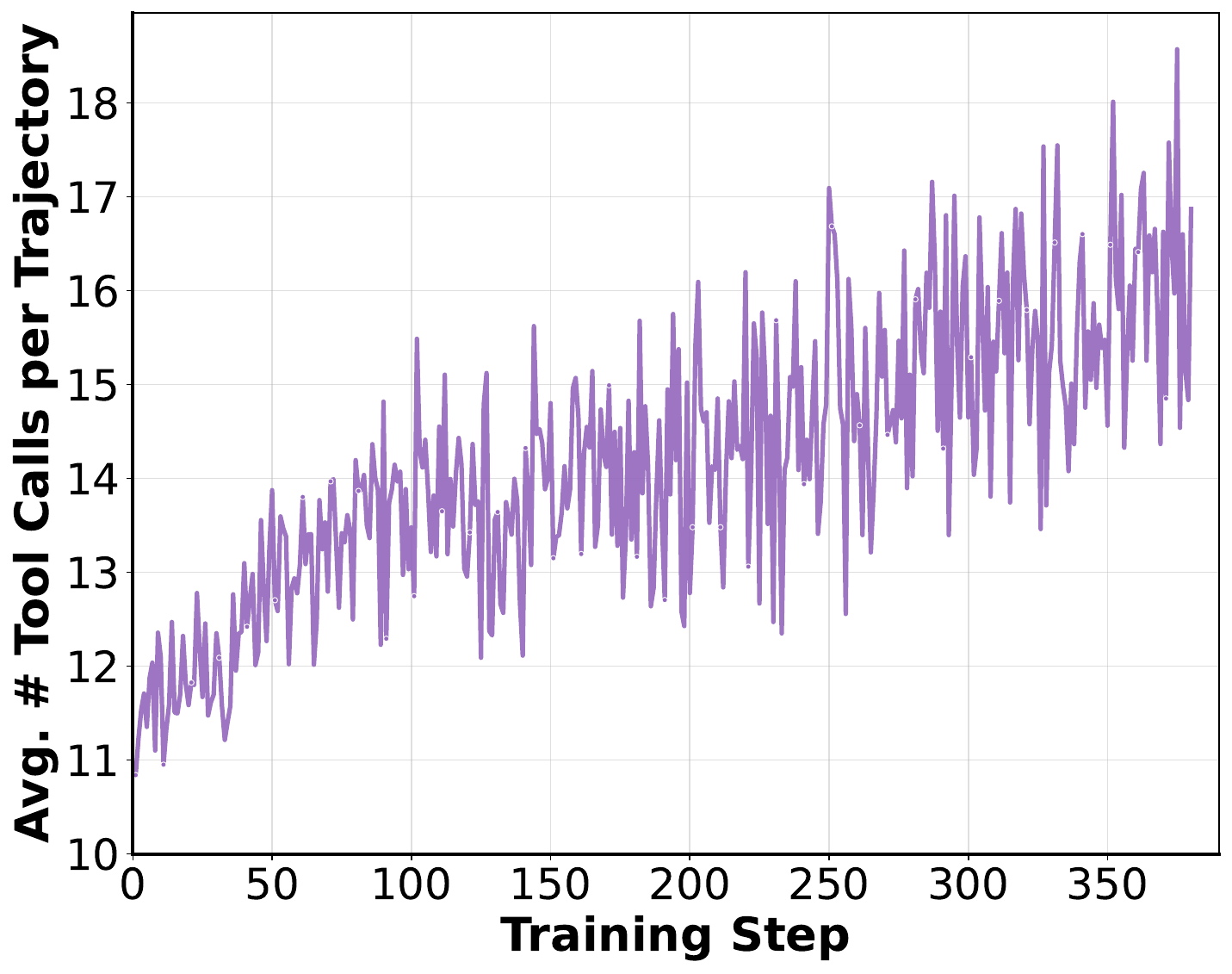}
\end{subfigure}
\hfill
\begin{subfigure}[t]{0.32\textwidth}  % 这里也用 [t]
    \centering
    \includegraphics[width=\textwidth]{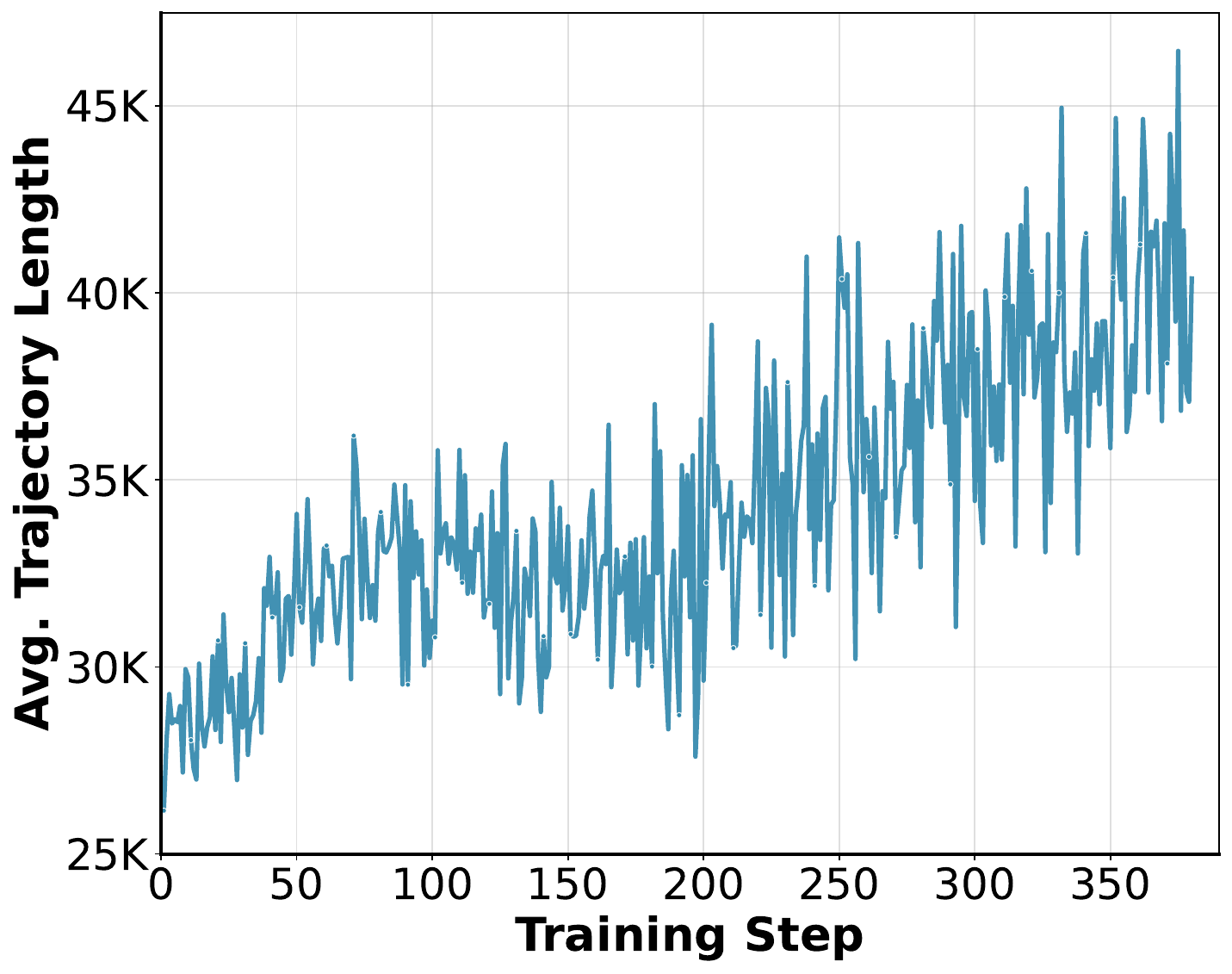}
\end{subfigure}
\hfill
\begin{subfigure}[t]{0.32\textwidth}  % 这里也用 [t]
    \centering
    \includegraphics[width=\textwidth]{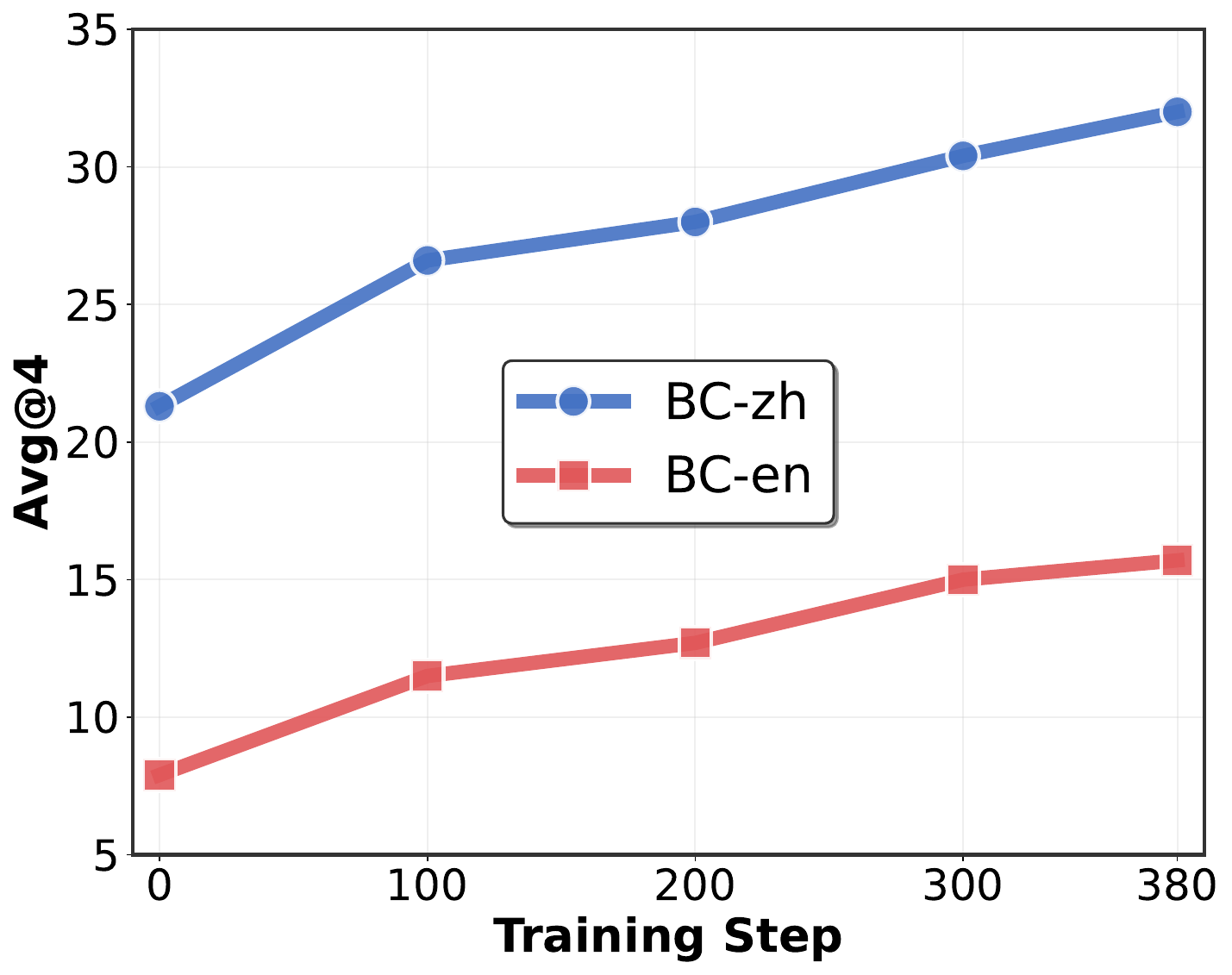}
\end{subfigure}
    \caption{\textbf{Left}: Average \# tool calls per trajectory during the RL training process. Each tool call (\texttt{search} or \texttt{browse}) is counted separately. \textbf{Middle}: Average trajectory length  (\# tokens) during the RL training process. 
    \textbf{Right}: The avg@4 scores of BrowseComp-en and BrowseComp-zh during the RL training process.}
% \caption{\textbf{Left}: Average # tool calls per trajectory during the RL training process. Each tool call (\texttt{search} or \texttt{browse}) is counted separately. \textbf{Middle}: Average trajectory length (number of tokens) during the RL training process. \textbf{Right}: The Avg@4 scores of BrowseComp-en and BrowseComp-zh during the RL training process.}

    \label{fig:scal-rl}
\end{figure*}

\subsection{Experimental Results}

% \jh{Please describe results categorized by different ``paragraph'' environements, not describing everything together.}
% As shown in Table~\ref{tab:model_comparison}, our \method{}-8B demonstrates strong performance across diverse information-seeking benchmarks. Following supervised fine-tuning, \method{}-8B (SFT) already exhibits competitive results, achieving 7.9 on BrowseComp, 72.6 on FRAMES, and 47.5 on XBench-DS. These results surpass the previous WebSailor-7B model across all evaluated benchmarks. The subsequent reinforcement learning phase yields substantial improvements, with our model achieving 15.7 on BrowseComp and 32.0 on BrowseComp-zh.

\noindent \textbf{\method{}-8B Establishes State-of-the-Art Performance at Its Scale} As shown in Table~\ref{tab:model_comparison}, our \method{}-8B demonstrates exceptional performance across diverse information-seeking benchmarks. Following supervised fine-tuning, \method{}-8B (SFT) already exhibits competitive results, surpassing WebSailor-7B with scores of 7.9\% on BrowseComp-en, 72.6\% on FRAMES, and 47.5\% on XBench-DS. The subsequent reinforcement learning phase yields substantial improvements, with our final model achieving 15.7\% on BrowseComp-en and 32.0\% on BrowseComp-zh. Notably, \method{}-8B (RL) surpasses the previous best sub-10B model, MiroThinker-8B-DPO-v0.1, across all benchmarks, establishing new state-of-the-art performance at its parameter scale.

% Despite having only 8B parameters, \method{}-8B consistently outperforms much larger open-source models. On both BrowseComp and BrowseComp-zh, our model surpasses WebSailor-72B and Kimi-K2 models. Similarly, on WebWalkerQA, we achieve 62.7, outperforming WebShaper-72B by 10 absolute points. Our model also demonstrates competitive performance on additional benchmarks, achieving 50.0 on GAIA and 75.7 on FRAMES. Overall, \method{}-8B establishes a new performance standard for parameter-efficient web agents.
% \jh{if we are going to mention our entity seeds are from browsecomp -- which I think we should describe -- then we should emphasize that such data actually generalize well. We can even mention we tried in-domain seeds as well from other benchs but they do not really help.}

\noindent \textbf{\method{}-8B Outperforms Much Larger Models} Despite having only 8B parameters, \method{}-8B consistently outperforms much larger open-source models, demonstrating remarkable parameter efficiency. On BrowseComp-en, BrowseComp-zh, FRAMES, and WebWalkerQA, our model achieves the best performance among open-source sub-100B models. Specifically, with scores of 15.7\% on BrowseComp-en, 32.0\% on BrowseComp-zh, and 75.7\% on FRAMES, our model surpasses these significantly larger models including WebSailor-72B and even larger model Kimi-K2. Similarly, on WebWalkerQA, we achieve 62.7\%, outperforming WebShaper-72B by 10 absolute points. Our model also demonstrates competitive performance on additional benchmarks, achieving 50.0\% on GAIA and 53.7\% on XBench-DS, rivaling models with more parameters like WebThinker-32B and WebSailor-32B.

% \jh{i think a more important aspect is not how it evolves, but where the seed entity comes from. We can emphasize the data are in different domains, it is not STEM}
\noindent \textbf{Strong Generalization from BrowseComp-Inspired Training Data} Notably, while our QA pair synthesis methodology, particularly the evolution stage, is inspired by BrowseComp-en, and the exemplars provided in synthesis prompts are both derived from BrowseComp-en, we ensure diversity by collecting seed entities from Wikipedia across different domains. The resulting model also generalizes effectively across diverse benchmarks and domains beyond the BrowseComp-en/zh. As demonstrated previously, \method{}-8B achieves strong performance across all information-seeking benchmarks, including GAIA, WebWalkerQA, FRAMES, and XBench-DS. This cross-benchmark success indicates substantial improvement in general information-seeking capabilities, demonstrating the generalization benefits of training on our challenging, complex queries. 
More remarkably, despite our training data not being STEM-focused, our model extends beyond information-seeking tasks. It achieves 17.3\% on the HLE benchmark, which encompasses questions from diverse academic disciplines.  This performance surpasses previous 32B models, including ASearcher-Web-QwQ and WebThinker-32B, further validating the robustness and transferability of our approach.

% \noindent \textbf{RL Training Dynamics and Performance Analysis} To analyze the training dynamics, we recorded the average \# tool calls per trajectory and the average response length (\# tokens) during RL training, with results presented in Figure~\ref{fig:scal-rl}. Throughout the RL training process, the average number of tool calls increases steadily from approximately 11 to over 16, significantly exceeding the typical range of fewer than 10 tool calls observed in previous work~\citep{gao2025turnsunlockinglonghorizonagentic}. This substantial increase suggests that our model learns to execute more sophisticated multi-step reasoning strategies compared to existing approaches. Concurrently, the average trajectory length grows during the RL process, scaling to over 40K tokens, as the \texttt{search} tool response maintain the majority source of trajectory length. Importantly, we observe that performance on both BrowseComp-en and BrowseComp-zh improves consistently throughout this process, like BrowseComp-en increases from 7.9 to 15.7, demonstrating a strong correlation between increased tool usage complexity and task performance. Similar to the emergence long CoT phenomenon observed in previous single-turn RL on math or logical tasks~\citep{guo2025deepseek,zeng2025simplerl,yu2025dapo,liu2025synlogic}, these findings validate the effectiveness of our RL training approach in developing more capable web agents that can tackle complex information-seeking tasks through extended reasoning chains.

\noindent \textbf{RL Training Dynamics and Performance Analysis} To analyze the training dynamics, we recorded the average number of tool calls per trajectory and the average response length (number of tokens) during RL training, with results presented in Figure~\ref{fig:scal-rl}. Throughout the RL training process, the average number of tool calls increases steadily from approximately 11 to over 16, significantly exceeding the typical range of fewer than 10 tool calls observed in previous work~\citep{gao2025turnsunlockinglonghorizonagentic}. This substantial increase suggests that our model learns to execute more sophisticated multi-step reasoning strategies. Concurrently, the average trajectory length grows during the RL process, scaling to over 40K tokens, as the \texttt{search} tool responses constitute the majority source of trajectory length. Importantly, we observe that performance on both BrowseComp-en and BrowseComp-zh improves consistently throughout this process, with BrowseComp-en increasing from 7.9\% to 15.7\%, demonstrating a strong correlation between increased tool usage complexity and task performance. Similar to the emergence of long chain-of-thought phenomena observed in previous single-turn RL studies on mathematical or logical tasks~\citep{guo2025deepseek,zeng2025simplerl,yu2025dapo,liu2025synlogic}, these findings validate the effectiveness of our RL training approach in developing advanced web agents that can tackle complex information-seeking tasks through extended reasoning chains.

% \jh{conclusion section?}

% These results validate the effectiveness of our training methodology, particularly the construction of challenging QA pairs and the progressive reinforcement learning approach. The substantial performance gains from RL training highlight the importance of our iterative refinement process in developing capable web agents.

% \subsection{Discussions}

\section{Conclusion}

In this work, we present \method{}, a query-answer data synthesis approach for training advanced web agents. By employing model-based exploration and iterative query evolution, we synthesize challenging query-answer pairs that require long-horizon reasoning spanning over 10 tool calling turns. Leveraging these high-quality QA pairs for supervised fine-tuning followed by GRPO-based reinforcement learning, our 8B model \method{}-8B achieves state-of-the-art performance across multiple information-seeking benchmarks, including BrowseComp-en, BrowseComp-zh, GAIA, WebWalkerQA, FRAMES, and XBench-DeepSearch. It establishes the best performance at its scale while significantly outperforming larger models such as WebSailor-72B.
Beyond information-seeking tasks, \method{}-8B demonstrates strong generalization on the academic HLE benchmark, outperforming previous 32B models by nearly 5 points. Our RL training enables the model to scale to over 16 tool calling turns on average, validating the effectiveness of our approach for training long-horizon web agents.
The success of \method{} demonstrates the potential of autonomously synthesizing challenging information-seeking QA pairs and leveraging supervised fine-tuning and reinforcement learning to build advanced, long-horizon web agents.

% \renewcommand\thefootnote{}\footnote{This is a numberless footnote.}

% \newpage
% \section*{Contributions}

% \noindent\textbf{Core Contributions}

% Yan Ma$*$$^{1,4}$, Linge Du$*$$^{1,3}$, Xuyang Shen$*$$\dagger$$^{1}$, $^\textrm{\Letter}$Junjie Yan$^{1}$ 
% \begingroup
% \renewcommand\thefootnote{}\footnote{$*$ Equal Contribution; $\dagger$ Project Lead; $^\textrm{\Letter}$ Corresponding Author}
% \addtocounter{footnote}{-1}
% \endgroup

% \noindent\textbf{Contributions}

% Shaoxiang Chen$^{1}$, Pengfei Li$^{1}$, Qibing Ren$^{1,2}$

% \noindent\textbf{Advisor}

% Junjie Yan$^{1}$, Pengfei Liu$^{2,4}$, Yuchao Dai$^{3}$, Lizhuang Ma$^{2}$

% \noindent\textbf{Affiliation}

% $^{1}$ MiniMax

% $^{2}$ Shanghai Jiao Tong University

% $^{3}$ Northwestern Polytechnical University

% $^{4}$ Generative Artificial Intelligence Lab (GAIR)

% \newpage
\bibliography{reference}

\begin{thebibliography}{32}
\providecommand{\natexlab}[1]{#1}
\providecommand{\url}[1]{\texttt{#1}}
\expandafter\ifx\csname urlstyle\endcsname\relax
  \providecommand{\doi}[1]{doi: #1}\else
  \providecommand{\doi}{doi: \begingroup \urlstyle{rm}\Url}\fi

\bibitem[{Anthropic}(2025)]{claude4_blog_2025}
{Anthropic}.
\newblock Introducing claude 4.
\newblock Technical report, 2025.
\newblock URL \url{https://www.anthropic.com/news/claude-4}.

\bibitem[Chen et~al.(2025)Chen, Ma, Zhuang, Nie, Zou, Liu, Green, Patel, Meng, Su, Sharifymoghaddam, Li, Hong, Shi, Liu, Thakur, Zhang, Gao, Chen, and Lin]{chen2025BrowseCompPlus}
Zijian Chen, Xueguang Ma, Shengyao Zhuang, Ping Nie, Kai Zou, Andrew Liu, Joshua Green, Kshama Patel, Ruoxi Meng, Mingyi Su, Sahel Sharifymoghaddam, Yanxi Li, Haoran Hong, Xinyu Shi, Xuye Liu, Nandan Thakur, Crystina Zhang, Luyu Gao, Wenhu Chen, and Jimmy Lin.
\newblock Browsecomp-plus: A more fair and transparent evaluation benchmark of deep-research agent.
\newblock \emph{arXiv preprint arXiv:2508.06600}, 2025.

\bibitem[Comanici et~al.(2025)Comanici, Bieber, Schaekermann, Pasupat, Sachdeva, Dhillon, Blistein, Ram, Zhang, Rosen, et~al.]{comanici2025gemini}
Gheorghe Comanici, Eric Bieber, Mike Schaekermann, Ice Pasupat, Noveen Sachdeva, Inderjit Dhillon, Marcel Blistein, Ori Ram, Dan Zhang, Evan Rosen, et~al.
\newblock Gemini 2.5: Pushing the frontier with advanced reasoning, multimodality, long context, and next generation agentic capabilities.
\newblock \emph{arXiv preprint arXiv:2507.06261}, 2025.

\bibitem[Gao et~al.(2025)Gao, Fu, Xie, Xu, He, Mei, Zhu, and Wu]{gao2025turnsunlockinglonghorizonagentic}
Jiaxuan Gao, Wei Fu, Minyang Xie, Shusheng Xu, Chuyi He, Zhiyu Mei, Banghua Zhu, and Yi~Wu.
\newblock Beyond ten turns: Unlocking long-horizon agentic search with large-scale asynchronous rl, 2025.
\newblock URL \url{https://arxiv.org/abs/2508.07976}.

\bibitem[Guo et~al.(2025)Guo, Yang, Zhang, Song, Zhang, Xu, Zhu, Ma, Wang, Bi, et~al.]{guo2025deepseek}
Daya Guo, Dejian Yang, Haowei Zhang, Junxiao Song, Ruoyu Zhang, Runxin Xu, Qihao Zhu, Shirong Ma, Peiyi Wang, Xiao Bi, et~al.
\newblock Deepseek-r1: Incentivizing reasoning capability in llms via reinforcement learning.
\newblock \emph{arXiv preprint arXiv:2501.12948}, 2025.

\bibitem[Jina.ai(2025)]{jina2025}
Jina.ai.
\newblock Jina, 2025.
\newblock URL \url{https://jina.ai}.

\bibitem[Krishna et~al.(2024)Krishna, Krishna, Mohananey, Schwarcz, Stambler, Upadhyay, and Faruqui]{krishna2024frames}
Satyapriya Krishna, Kalpesh Krishna, Anhad Mohananey, Steven Schwarcz, Adam Stambler, Shyam Upadhyay, and Manaal Faruqui.
\newblock Fact, fetch, and reason: A unified evaluation of retrieval-augmented generation.
\newblock In \emph{North American Chapter of the Association for Computational Linguistics}, 2024.
\newblock URL \url{https://api.semanticscholar.org/CorpusID:272753636}.

\bibitem[Li et~al.(2025{\natexlab{a}})Li, Zhang, Yin, Zhang, Ou, Wu, Yin, Li, Tao, Wang, et~al.]{li2025websailor}
Kuan Li, Zhongwang Zhang, Huifeng Yin, Liwen Zhang, Litu Ou, Jialong Wu, Wenbiao Yin, Baixuan Li, Zhengwei Tao, Xinyu Wang, et~al.
\newblock Websailor: Navigating super-human reasoning for web agent.
\newblock \emph{arXiv preprint arXiv:2507.02592}, 2025{\natexlab{a}}.

\bibitem[Li et~al.(2025{\natexlab{b}})Li, Lin, Jiang, Cao, Liu, Zhang, Huang, Chen, Sun, Wang, Lu, Qin, Zhu, Yao, Fan, Li, Wang, Liu, Zhu, Zhu, Shi, Wang, Guan, Tang, Liu, Jiang, Yang, Liu, Zhang, and Zhou]{li2025chainofagentsendtoendagentfoundation}
Weizhen Li, Jianbo Lin, Zhuosong Jiang, Jingyi Cao, Xinpeng Liu, Jiayu Zhang, Zhenqiang Huang, Qianben Chen, Weichen Sun, Qiexiang Wang, Hongxuan Lu, Tianrui Qin, Chenghao Zhu, Yi~Yao, Shuying Fan, Xiaowan Li, Tiannan Wang, Pai Liu, King Zhu, He~Zhu, Dingfeng Shi, Piaohong Wang, Yeyi Guan, Xiangru Tang, Minghao Liu, Yuchen~Eleanor Jiang, Jian Yang, Jiaheng Liu, Ge~Zhang, and Wangchunshu Zhou.
\newblock Chain-of-agents: End-to-end agent foundation models via multi-agent distillation and agentic rl, 2025{\natexlab{b}}.
\newblock URL \url{https://arxiv.org/abs/2508.13167}.

\bibitem[Li et~al.(2025{\natexlab{c}})Li, Jin, Dong, Qian, Zhu, Wu, Wen, and Dou]{li2025webthinker}
Xiaoxi Li, Jiajie Jin, Guanting Dong, Hongjin Qian, Yutao Zhu, Yongkang Wu, Ji-Rong Wen, and Zhicheng Dou.
\newblock Webthinker: Empowering large reasoning models with deep research capability.
\newblock \emph{arXiv preprint arXiv:2504.21776}, 2025{\natexlab{c}}.

\bibitem[Liu et~al.(2024)Liu, Feng, Xue, Wang, Wu, Lu, Zhao, Deng, Zhang, Ruan, et~al.]{liu2024deepseek}
Aixin Liu, Bei Feng, Bing Xue, Bingxuan Wang, Bochao Wu, Chengda Lu, Chenggang Zhao, Chengqi Deng, Chenyu Zhang, Chong Ruan, et~al.
\newblock Deepseek-v3 technical report.
\newblock \emph{arXiv preprint arXiv:2412.19437}, 2024.

\bibitem[Liu et~al.(2025)Liu, Fan, Jiang, Ding, Hu, Zhang, Shi, Weng, Chen, Chen, et~al.]{liu2025synlogic}
Junteng Liu, Yuanxiang Fan, Zhuo Jiang, Han Ding, Yongyi Hu, Chi Zhang, Yiqi Shi, Shitong Weng, Aili Chen, Shiqi Chen, et~al.
\newblock Synlogic: Synthesizing verifiable reasoning data at scale for learning logical reasoning and beyond.
\newblock \emph{arXiv preprint arXiv:2505.19641}, 2025.

\bibitem[Liu et~al.(2023)Liu, Yang, Huang, Zhang, Huang, Wei, Deng, Sun, and Zhang]{liu2023calibrating}
Yuxuan Liu, Tianchi Yang, Shaohan Huang, Zihan Zhang, Haizhen Huang, Furu Wei, Weiwei Deng, Feng Sun, and Qi~Zhang.
\newblock Calibrating llm-based evaluator.
\newblock \emph{arXiv preprint arXiv:2309.13308}, 2023.

\bibitem[Mialon et~al.(2024)Mialon, Fourrier, Wolf, LeCun, and Scialom]{mialon2024gaia}
Gr{\'e}goire Mialon, Cl{\'e}mentine Fourrier, Thomas Wolf, Yann LeCun, and Thomas Scialom.
\newblock {GAIA}: a benchmark for general {AI} assistants.
\newblock In \emph{The Twelfth International Conference on Learning Representations}, 2024.
\newblock URL \url{https://openreview.net/forum?id=fibxvahvs3}.

\bibitem[{MiroMind Team}(2025)]{miromind2025odr}
{MiroMind Team}.
\newblock Miromind open deep research, 2025.
\newblock URL \url{https://miromind.ai/blog/miromind-open-deep-research}.
\newblock Blog post, August 2025.

\bibitem[{OpenAI}(2025)]{openai_deep_research_2025}
{OpenAI}.
\newblock Introducing deep research.
\newblock Technical report, 2025.
\newblock URL \url{https://openai.com/index/introducing-deep-research/}.

\bibitem[Pahuja et~al.(2025)Pahuja, Lu, Rosset, Gou, Mitra, Whitehead, Su, and Awadallah]{pahuja-etal-2025-explorer}
Vardaan Pahuja, Yadong Lu, Corby Rosset, Boyu Gou, Arindam Mitra, Spencer Whitehead, Yu~Su, and Ahmed~Hassan Awadallah.
\newblock Explorer: Scaling exploration-driven web trajectory synthesis for multimodal web agents.
\newblock In \emph{Findings of the Association for Computational Linguistics: ACL 2025}, pages 6300--6323, Vienna, Austria, July 2025. Association for Computational Linguistics.
\newblock ISBN 979-8-89176-256-5.
\newblock URL \url{https://aclanthology.org/2025.findings-acl.326/}.

\bibitem[{Perplexity AI}(2025)]{perplexity_deep_research_2025}
{Perplexity AI}.
\newblock Introducing perplexity deep research.
\newblock \url{https://www.perplexity.ai/hub/blog/introducing-perplexity-deep-research}, February 2025.
\newblock Accessed: 2025-07-25.

\bibitem[Phan et~al.(2025)Phan, Gatti, Han, Li, Hu, Zhang, Zhang, Shaaban, Ling, Shi, et~al.]{phan2025humanity}
Long Phan, Alice Gatti, Ziwen Han, Nathaniel Li, Josephina Hu, Hugh Zhang, Chen Bo~Calvin Zhang, Mohamed Shaaban, John Ling, Sean Shi, et~al.
\newblock Humanity's last exam.
\newblock \emph{arXiv preprint arXiv:2501.14249}, 2025.

\bibitem[Shao et~al.(2024)Shao, Wang, Zhu, Xu, Song, Bi, Zhang, Zhang, Li, Wu, et~al.]{shao2024deepseekmath}
Zhihong Shao, Peiyi Wang, Qihao Zhu, Runxin Xu, Junxiao Song, Xiao Bi, Haowei Zhang, Mingchuan Zhang, YK~Li, Yang Wu, et~al.
\newblock Deepseekmath: Pushing the limits of mathematical reasoning in open language models.
\newblock \emph{arXiv preprint arXiv:2402.03300}, 2024.

\bibitem[Tao et~al.(2025)Tao, Wu, Yin, Zhang, Li, Shen, Li, Zhang, Wang, Jiang, et~al.]{tao2025webshaper}
Zhengwei Tao, Jialong Wu, Wenbiao Yin, Junkai Zhang, Baixuan Li, Haiyang Shen, Kuan Li, Liwen Zhang, Xinyu Wang, Yong Jiang, et~al.
\newblock Webshaper: Agentically data synthesizing via information-seeking formalization.
\newblock \emph{arXiv preprint arXiv:2507.15061}, 2025.

\bibitem[Team et~al.(2025)Team, Bai, Bao, Chen, Chen, Chen, Chen, Chen, Chen, Chen, et~al.]{team2025kimi}
Kimi Team, Yifan Bai, Yiping Bao, Guanduo Chen, Jiahao Chen, Ningxin Chen, Ruijue Chen, Yanru Chen, Yuankun Chen, Yutian Chen, et~al.
\newblock Kimi k2: Open agentic intelligence.
\newblock \emph{arXiv preprint arXiv:2507.20534}, 2025.

\bibitem[Wei et~al.(2025)Wei, Sun, Papay, McKinney, Han, Fulford, Chung, Passos, Fedus, and Glaese]{wei2025browsecomp}
Jason Wei, Zhiqing Sun, Spencer Papay, Scott McKinney, Jeffrey Han, Isa Fulford, Hyung~Won Chung, Alex~Tachard Passos, William Fedus, and Amelia Glaese.
\newblock Browsecomp: A simple yet challenging benchmark for browsing agents.
\newblock \emph{arXiv preprint arXiv:2504.12516}, 2025.

\bibitem[Wu et~al.(2025{\natexlab{a}})Wu, Li, Fang, Yin, Zhang, Tao, Zhang, Xi, Fu, Jiang, et~al.]{wu2025webdancer}
Jialong Wu, Baixuan Li, Runnan Fang, Wenbiao Yin, Liwen Zhang, Zhengwei Tao, Dingchu Zhang, Zekun Xi, Gang Fu, Yong Jiang, et~al.
\newblock Webdancer: Towards autonomous information seeking agency.
\newblock \emph{arXiv preprint arXiv:2505.22648}, 2025{\natexlab{a}}.

\bibitem[Wu et~al.(2025{\natexlab{b}})Wu, Yin, Jiang, Wang, Xi, Fang, Zhou, Xie, and Huang]{wu2025webwalker}
Jialong Wu, Wenbiao Yin, Yong Jiang, Zhenglin Wang, Zekun Xi, Runnan Fang, Deyu Zhou, Pengjun Xie, and Fei Huang.
\newblock Webwalker: Benchmarking llms in web traversal.
\newblock \emph{ArXiv}, abs/2501.07572, 2025{\natexlab{b}}.
\newblock URL \url{https://api.semanticscholar.org/CorpusID:275471576}.

\bibitem[Xbench-Team(2025)]{xbench2025}
Xbench-Team.
\newblock Xbench-deepsearch, 2025.
\newblock URL \url{https://xbench.org/agi/aisearch}.

\bibitem[Yang et~al.(2025)Yang, Li, Yang, Zhang, Hui, Zheng, Yu, Gao, Huang, Lv, et~al.]{yang2025qwen3}
An~Yang, Anfeng Li, Baosong Yang, Beichen Zhang, Binyuan Hui, Bo~Zheng, Bowen Yu, Chang Gao, Chengen Huang, Chenxu Lv, et~al.
\newblock Qwen3 technical report.
\newblock \emph{arXiv preprint arXiv:2505.09388}, 2025.

\bibitem[Yao et~al.(2023)Yao, Zhao, Yu, Du, Shafran, Narasimhan, and Cao]{yao2023react}
Shunyu Yao, Jeffrey Zhao, Dian Yu, Nan Du, Izhak Shafran, Karthik Narasimhan, and Yuan Cao.
\newblock React: Synergizing reasoning and acting in language models.
\newblock In \emph{International Conference on Learning Representations (ICLR)}, 2023.

\bibitem[Yu et~al.(2025)Yu, Zhang, Zhu, Yuan, Zuo, Yue, Dai, Fan, Liu, Liu, et~al.]{yu2025dapo}
Qiying Yu, Zheng Zhang, Ruofei Zhu, Yufeng Yuan, Xiaochen Zuo, Yu~Yue, Weinan Dai, Tiantian Fan, Gaohong Liu, Lingjun Liu, et~al.
\newblock Dapo: An open-source llm reinforcement learning system at scale.
\newblock \emph{arXiv preprint arXiv:2503.14476}, 2025.

\bibitem[Zeng et~al.(2025)Zeng, Huang, Liu, Liu, He, Ma, and He]{zeng2025simplerl}
Weihao Zeng, Yuzhen Huang, Qian Liu, Wei Liu, Keqing He, Zejun Ma, and Junxian He.
\newblock Simplerl-zoo: Investigating and taming zero reinforcement learning for open base models in the wild.
\newblock \emph{arXiv preprint arXiv:2503.18892}, 2025.

\bibitem[{Zhipu AI}(2025)]{glm45_blog_2025}
{Zhipu AI}.
\newblock {GLM-4.5: Reasoning, Coding, and Agentic Abililties}.
\newblock \url{https://z.ai/blog/glm-4.5}, July 2025.
\newblock Accessed: 2025-08-07.

\bibitem[Zhou et~al.(2025)Zhou, Leon, Ying, Zhang, Shao, Ye, Chong, Jin, Xie, Cao, et~al.]{zhou2025browsecomp}
Peilin Zhou, Bruce Leon, Xiang Ying, Can Zhang, Yifan Shao, Qichen Ye, Dading Chong, Zhiling Jin, Chenxuan Xie, Meng Cao, et~al.
\newblock Browsecomp-zh: Benchmarking web browsing ability of large language models in chinese.
\newblock \emph{arXiv preprint arXiv:2504.19314}, 2025.

\end{thebibliography}

\appendix
\section{Trajectory Example}

% Here, we provide a trajectory example to show our agent framework.
To illustrate how our agent framework operates in practice, we provide a detailed trajectory example from the \method{} supervised fine-tuning data. This example demonstrates the agent's multi-step reasoning process, thought, tool call, and tool response when handling complex queries.

\label{app:traj-example}

\begin{outerframe}

% \begin{topbanner}
% CASE TRAJECTORY OF AFM ON GAIA.
% \end{topbanner}

\medskip

\textbf{Question:} In a county adjacent to the nation's capital, a small press gaming company co-founded by someone whose previous pulp adventure game had won a silver-level award in a category focusing on game mechanics excellence. This person reached a specific age milestone during an online funding campaign spanning consecutive years that occurred exactly 6 years after their game's award. What was the exact final funding amount raised by this campaign, expressed in US dollars without the dollar sign or comma separators?
\medskip

\begin{turnbarline}
Step 1
\end{turnbarline}

\begin{thoughtbox}
\texttt{\textless think\textgreater}

This is a complex multi-part query that I need to break down systematically. Let me identify the key components:

1. A county adjacent to the nation's capital (so near Washington DC)

\ldots

% This sounds like it could be referring to a Kickstarter or similar crowdfunding campaign. The "silver-level award" and "game mechanics excellence" suggests this might be an ENnie Award (silver ENnie) or similar gaming industry award.

Let me start by searching for information about gaming companies near Washington DC that had crowdfunding campaigns, and awards for pulp adventure games.

\texttt{\textless/think\textgreater}

I need to work through this step-by-step to identify the specific gaming company, award, and crowdfunding campaign. Let me start by searching for information about gaming companies near Washington DC and awards for pulp adventure games.

\end{thoughtbox}

\begin{actionbox}
% \% hi
\texttt{\textless tool\_call\textgreater}

\{``name'': ``\texttt{search}'', ``arguments'': \{``queries'': [``small press gaming company Washington DC area'', ``pulp adventure game silver award game mechanics ENnie'', ``gaming company county adjacent Washington DC crowdfunding'']\}\}

\texttt{\textless/tool\_call\textgreater}
\end{actionbox}

\begin{obbox}
\texttt{\textless tool\_response\textgreater}

\medskip

--- search result for [small press gaming company Washington DC area] ---

\textless title\textgreater op Washington, DC Gaming Companies (10) - Built In \textless/title\textgreater

\textless url\textgreater https://builtin.com/companies/location/washington-dc/type/gaming-companies \textless/url\textgreater

\textless snippet\textgreater
Top Washington, DC Gaming Companies (10) · Edge Systems Technologies · Something Wicked Games · Polygon · Implicit Solution · BetMGM · Snaapiq · Microcoaching.
\textless/snippet\textgreater

% \textless title\textgreater Wood-Aged Vernor's Ginger Ale Was `Deliciously Different' \textless/title\textgreater

% \textless url\textgreater http://www.metnews.com/articles/2005/reminiscing101305.htm \textless/url\textgreater

% \textless snippet\textgreater
% It was a mix of 19 ingredients, including ginger, vanilla and natural flavorings. When Vernor was called off to war in 1862, he stored the \ldots
% \textless/snippet\textgreater

\ldots

\texttt{\textless/tool\_response\textgreater}
\end{obbox}

\end{outerframe}

\begin{outerframe}

\begin{turnbarline}
Step 2
\end{turnbarline}

\begin{thoughtbox}
\texttt{\textless think\textgreater}

From the search results, I can see some relevant information:

1. I found mention of ``Spirit of the Century'' which won a Silver ENnie for Best Rules in 2007, and it's described as ``A really well put together game of pulp adventure''

\ldots

\texttt{\textless/think\textgreater}

Let me search more specifically for information about Spirit of the Century, which appears to be a pulp adventure game that won a silver ENnie, and its publisher.

\end{thoughtbox}

\begin{actionbox}
% \% hi
\texttt{\textless tool\_call\textgreater}

\{``name'': ``search'', ``arguments'': \{``queries'': [``Spirit of the Century silver ENnie 2007 Evil Hat Productions'', ``Evil Hat Productions Maryland Virginia crowdfunding Kickstarter'', ``Fred Hicks Evil Hat Productions age milestone funding campaign'']\}\}

\texttt{\textless/tool\_call\textgreater}
\end{actionbox}

\begin{obbox}
\texttt{\textless tool\_response\textgreater}

\medskip

--- search result for [Spirit of the Century silver ENnie 2007 Evil Hat Productions] ---

\textless title\textgreater Spirit of the Century - Wikipedia
\textless/title\textgreater

\textless url\textgreater https://en.wikipedia.org/wiki/Spirit\_of\_the\_Century
\textless/url\textgreater

\textless snippet\textgreater
Spirit of the Century is a pulp role-playing game published by Evil Hat Productions, and based on Evil Hat's FATE system. \ldots 2007 Silver ENnie Award for Best \ldots
\textless/snippet\textgreater

% \textless title\textgreater Wood-Aged Vernor's Ginger Ale Was `Deliciously Different' \textless/title\textgreater

% \textless url\textgreater http://www.metnews.com/articles/2005/reminiscing101305.htm \textless/url\textgreater

% \textless snippet\textgreater
% It was a mix of 19 ingredients, including ginger, vanilla and natural flavorings. When Vernor was called off to war in 1862, he stored the \ldots
% \textless/snippet\textgreater

\ldots

\texttt{\textless/tool\_response\textgreater}
\end{obbox}

\begin{turnbarline}
    STEP 3
\end{turnbarline}

\begin{thoughtbox}
\texttt{\textless think\textgreater}

Great! Now I have some key information:

1. Spirit of the Century is a pulp adventure game by Evil Hat Productions

\ldots
% 2. It won a Silver ENnie Award for Best Rules in 2007
% 3. Evil Hat Productions is co-founded by Fred Hicks (and Rob Donoghue)
% 4. Evil Hat is based in Silver Spring, Maryland (from the Washington Post article mention)
% 5. Silver Spring, Maryland is indeed in a county adjacent to Washington DC (Montgomery County)

% Now I need to find information about:
% - Fred Hicks reaching an age milestone during a funding campaign
% - A campaign that spanned consecutive years
% - A campaign that occurred exactly 6 years after the 2007 award (so 2013)
% - The exact funding amount

% Let me search for Evil Hat's crowdfunding campaigns around 2013, and specifically look for information about Fred Hicks' age milestones.

\texttt{\textless/think\textgreater}

Perfect! I found that Spirit of the Century by Evil Hat Productions won a Silver ENnie for Best Rules in 2007, and Evil Hat is based in Silver Spring, Maryland (Montgomery County, adjacent to DC). Now let me search for their crowdfunding campaigns around 2013 (6 years after 2007) and information about Fred Hicks' age milestones.

\end{thoughtbox}

\end{outerframe}

\begin{outerframe}

\begin{actionbox}
% \% hi
\texttt{\textless tool\_call\textgreater}

\{``name'': ``search'', ``arguments'': \{``queries'': [``Evil Hat Productions Kickstarter 2013 2014 Fred Hicks'', ``Fate Core Kickstarter 2013 funding amount final total'', ``Fred Hicks 40th birthday milestone age Fate Core campaign'']\}\}

\texttt{\textless/tool\_call\textgreater}
\end{actionbox}

\begin{obbox}
\texttt{\textless tool\_response\textgreater}

\medskip

--- search result for [Evil Hat Productions Kickstarter 2013 2014 Fred Hicks] ---

\textless title\textgreater Race to Adventure! A Spirit of the Century™ Board Game - Kickstarter
\textless/title\textgreater

\textless url\textgreater https://www.kickstarter.com/projects/evilhat/race-to-adventure-a-spirit-of-the-centurytm-board/posts?ref=project\_share
\textless/url\textgreater

\textless snippet\textgreater
Evil Hat Productions is raising funds for Race to Adventure! A Spirit of the Century™ Board Game on Kickstarter! Race to Adventure!

\textless/snippet\textgreater

% \textless title\textgreater Wood-Aged Vernor's Ginger Ale Was `Deliciously Different' \textless/title\textgreater

% \textless url\textgreater http://www.metnews.com/articles/2005/reminiscing101305.htm \textless/url\textgreater

% \textless snippet\textgreater
% It was a mix of 19 ingredients, including ginger, vanilla and natural flavorings. When Vernor was called off to war in 1862, he stored the \ldots
% \textless/snippet\textgreater

\ldots

\texttt{\textless/tool\_response\textgreater}
\end{obbox}

\begin{turnbarline}
    STEP 4
\end{turnbarline}

\begin{thoughtbox}
\texttt{\textless think\textgreater}

I found some crucial information! The Fate Core Kickstarter by Evil Hat Productions raised exactly \$433,365 and ended on January 30, 2013. This is 6 years after the 2007 ENnie award for Spirit of the Century.

\ldots

\texttt{\textless/think\textgreater}

Excellent! I found that the Fate Core Kickstarter raised \$433,365 and ended on January 30, 2013 - which is exactly 6 years after the 2007 ENnie award. Let me get more details about this campaign to confirm it spans consecutive years and involves an age milestone.

\end{thoughtbox}

\begin{actionbox}
% \% hi
\texttt{\textless tool\_call\textgreater}

\{``name'': ``browse'', ``arguments'': \{

``url'':
``https://www.kickstarter.com/projects/evilhat/fate-core'',

``query'': When did this campaign start and end? Did it span 2012 and 2013? Any mention of Fred Hicks birthday or age milestone during the campaign?\}\}

\texttt{\textless/tool\_call\textgreater}
\end{actionbox}

\end{outerframe}

\begin{outerframe}

\begin{obbox}
\texttt{\textless tool\_response\textgreater}

\medskip

The campaign started on **December 3, 2012**, and ended on **January 29, 2013**, spanning **56 days** and crossing over from **2012 to 2013**.  

There is **no mention of Fred Hicks' birthday or age milestone** in the provided content.

\texttt{\textless/tool\_response\textgreater}
\end{obbox}

\begin{turnbarline}
    STEP 5
\end{turnbarline}

\begin{thoughtbox}
\texttt{\textless think\textgreater}

Perfect! This confirms that the Fate Core Kickstarter:

\ldots

\texttt{\textless/think\textgreater}

Perfect! The Fate Core campaign spanned consecutive years (December 3, 2012 to January 29, 2013) and occurred exactly 6 years after the 2007 ENnie. Now let me search for information about Fred Hicks' birthday and age milestones during this period.

\end{thoughtbox}

\begin{actionbox}
% \% hi
\texttt{\textless tool\_call\textgreater}

\{``name'': ``search'', ``arguments'': \{``queries'': [``Fred Hicks birthday December January 40 years old'', ``Fred Hicks born 1972 1973 Evil Hat'', ``Fred Hicks 40th birthday December 2012 January 2013'']\}\}

\texttt{\textless/tool\_call\textgreater}
\end{actionbox}

\begin{obbox}
\texttt{\textless tool\_response\textgreater}

\medskip

--- search result for [Fred Hicks birthday December January 40 years old] ---

\textless title\textgreater Happy birthday to the first general manager of Forsythe Associates ...
\textless/title\textgreater

\textless url\textgreater https://www.facebook.com/groups/649530655242462/posts/1496564073872445
\textless/url\textgreater

\textless snippet\textgreater
Happy birthday to the first general manager of Forsythe Associates [1962-1971], Fred Hicks [1920-1971].
\textless/snippet\textgreater

% \textless title\textgreater Wood-Aged Vernor's Ginger Ale Was `Deliciously Different' \textless/title\textgreater

% \textless url\textgreater http://www.metnews.com/articles/2005/reminiscing101305.htm \textless/url\textgreater

% \textless snippet\textgreater
% It was a mix of 19 ingredients, including ginger, vanilla and natural flavorings. When Vernor was called off to war in 1862, he stored the \ldots
% \textless/snippet\textgreater

\ldots

\texttt{\textless/tool\_response\textgreater}
\end{obbox}

\begin{turnbarline}
    STEP 6
\end{turnbarline}

\begin{thoughtbox}
\texttt{\textless think\textgreater}

Perfect! I found the crucial information. \ldots

\texttt{\textless/think\textgreater}

% Perfect! I found the key missing piece. According to BoardGameGeek, Fred Hicks was born on **January 13, 1972**.

% Let me verify all the details match:
Therefore, the exact final funding amount raised by this campaign, expressed in US dollars without the dollar sign or comma separators, is:
**433365**
\end{thoughtbox}

\end{outerframe}

\section{Details of \method{}-QA Synthesis Framework}
\label{app-prompt-details}
Compared with previous works~\citep{li2025websailor,tao2025webshaper} involving complex synthesis processes, our approach presents a simpler prompting-based framework. Here, we provide details about the synthesis methodology to facilitate reproducibility.

Our framework operates through two stages, each guided by carefully designed prompting strategies that leverage large language models for autonomous information gathering and query evolution.

\subsection{Model-Based Exploration}

We first collect a large number of entities from Wikipedia.  In the first stage, we provide a seed entity as the search entry point along with three exemplar QA pairs from BrowseComp-en to demonstrate the desired question characteristics. The prompting strategy encourages the model to conduct search and browsing activities starting from the seed entity, then synthesize a challenging query-answer pair using the collected knowledge.

The key instruction emphasizes creating challenging queries with subtle and obscured clues. We explicitly prompt the model to ensure that while the question should be challenging, the answer must remain unique and verifiable through the information space it constructs. This stage results in initial QA pairs that incorporate multi-website reasoning.

\begin{tcolorbox}[colback=gray!10, colframe=black, title=Model-Based Exploration Prompt, breakable]
\small
\texttt{You need to create a challenging question for deep search based on real information.}
\\\\
\texttt{You should start by collecting information from the internet, then select a truth, and create a question where the truth needs to be discovered through search.}\\
\texttt{You will start with a random "seed", then search and browse for whatever you want on the Internet, and create the question and truth from the information you gather.}\\
\texttt{You should provide several subtle and blurred clues to make the question challenging, while ensuring the truth is unique.}
\\\\
\texttt{There are some examples:}\\
\texttt{\{examples\}}
\\\\
\texttt{Let's start, with the seed of "\{seed\}".}\\
\texttt{You need to provide the following information in the final <answer></answer> tag:}\\
\texttt{<question>}\\
\texttt{\{\{The challenging question you created based on real information.\}\}}\\
\texttt{</question>}\\
\texttt{<truth>}\\
\texttt{\{\{The one and only exact truth to the question.\}\}}\\
\texttt{</truth>}
\\\\
\texttt{IMPORTANT: You must include the <question> and <truth> tags in your final response for the system to parse your answer correctly. Do not provide any other response format.}
\end{tcolorbox}

\subsection{Iterative Query Evolution}

We provide the full trajectory from the first stage as input to the evolution process. The second stage systematically transforms the initial QA pair into a more challenging variant through strategic information reduction and obfuscation. The prompting strategy provides the model with the original question-answer pair and explicit instructions for three primary evolution mechanisms: (1) removing redundant or overly explicit descriptions that provide multiple pathways to the answer, (2) systematically obfuscating specific details such as dates, locations, and proper names with vaguer descriptors, and (3) searching for alternative terminologies to replace explicit entity mentions.
The evolution process can iterate up to five times, with each cycle potentially increasing reasoning complexity. 

This prompting-based approach eliminates the need for complex graph construction heuristics or predefined evolution rules, instead leveraging the model's natural language understanding and web exploration capabilities. The framework's simplicity enables easy and quick reproducibility for generating high-quality challenging QA pairs.

\begin{tcolorbox}[colback=blue!10, colframe=blue!50!black, title=Iterative Query Evolution Prompt, breakable]
\small
\texttt{You need to make the following question more challenging while keeping the truth unique.}
\\\\
\texttt{Original question: \{question\}}\\
\texttt{Original truth: \{answer\}}
\\\\
\texttt{You should make the question more challenging in the following ways:}\\
\texttt{1. Remove some descriptions, especially when there are multiple descriptions that can lead to the truth answer in the question}\\
\texttt{2. Make one description more vague (such as date, location, name, etc.) in the question while keeping the truth answer unique}\\
\texttt{3. Search for new descriptions or alternative terms to replace specific entities in the question}
\\\\
\texttt{You can use search and browse tools in this process. Make sure the improved question is more challenging but the truth remains unique.}
\\\\
\texttt{You can iteratively make the question more and more challenging using these approaches up to 5 times. For each iteration, provide the evolved question within <question> and </question> tags. Provide the final improved question and truth pair in the final <answer></answer> tag:}\\
\texttt{<question>}\\
\texttt{\{\{The improved, more challenging and complex question.\}\}}\\
\texttt{</question>}\\
\texttt{<truth>}\\
\texttt{\{\{The same exact truth.\}\}}\\
\texttt{</truth>}
\\\\
\texttt{IMPORTANT: You must include the <question> and <truth> tags in your final response for the system to parse your answer correctly. Do not provide any other response format.}
\end{tcolorbox}

\section{Tool Schemas}
\label{sec:tool-schemas}

Our system utilizes two primary tools for web-based information retrieval: a search tool for discovering relevant content and a browse tool for detailed content analysis. The following sections detail the schemas for each tool.

\subsection{Search Tool}

The search tool enables web search functionality through Google search API, allowing the model to discover relevant URLs and content snippets based on input queries.

\begin{tcolorbox}[colback=gray!5!white,colframe=gray!75!black,title=Search Tool Schema,  boxrule=1pt]
\begin{lstlisting}[style=noframe]
tool_schema:
  type: function
  function:
    name: search
    description: Web search.
    parameters:
      type: object
      properties:
        queries:
          type: array
          description: The queries will be sent to Google. You will 
                      get the brief search results with (title, url, 
                      snippet)s for each query.
          items:
            type: string
      required:
        - queries
\end{lstlisting}
\end{tcolorbox}

The search tool accepts an array of query strings and returns search results containing titles, URLs, and content snippets for each query. It is configured to return the top 10 results per query.

\subsection{Browse Tool}

The browse tool provides detailed content analysis capabilities by fetching and processing content from specific URLs using a combination of content extraction and language model analysis.

% \begin{tcolorbox}[colback=blue!5!white,colframe=blue!75!black,title=Browse Tool Schema]
% \begin{lstlisting}[basicstyle=\footnotesize\ttfamily]
% tool_schema:
%   type: function
%   function:
%     name: browse
%     description: Explore specific information in a url.
%     parameters:
%       type: object
%       properties:
%         url:
%           type: string
%           description: The url will be browsed, and the content 
%                       will be sent to a Large Language Model (LLM) 
%                       as the based information to answer a query.
%         query:
%           type: string
%           description: The query to this url content. You will 
%                       get an answer by another LLM.
%       required:
%         - url
%         - query
% \end{lstlisting}
% \end{tcolorbox}

\begin{tcolorbox}[
  title=Browse Tool Schema,
  colback=blue!5!white,
  colframe=blue!75!black,
  boxrule=1pt,
  % rounded,
  % sharp corners=northwest,
  % breakable
]
\begin{lstlisting}[style=noframe]
tool_schema:
  type: function
  function:
    name: browse
    description: Explore specific information in a url.
    parameters:
      type: object
      properties:
        url:
          type: string
          description: The url will be browsed, and the content 
                      will be sent to a Large Language Model (LLM) 
                      as the based information to answer a query.
        query:
          type: string
          description: The query to this url content. You will 
                      get an answer by another LLM.
      required:
        - url
        - query
\end{lstlisting}
\end{tcolorbox}

The browse tool takes a URL and a specific query as input. It uses Jina~\citep{jina2025} for content extraction and Gemini Flash~\citep{comanici2025gemini} as the generation engine to analyze the retrieved content and provide targeted answers to the input query. 

% \section{Appendix / supplemental material}

% \input{src/appendix1}
% \input{src/appendix2}
\end{document}